%% file: ijcai23.tex
\documentclass{article}
\usepackage{ijcai23}

\usepackage{times}
\usepackage{soul}
\usepackage{url}
\usepackage[hidelinks]{hyperref}
\usepackage[utf8]{inputenc}
\usepackage[small]{caption}
\usepackage{graphicx}
\usepackage{subcaption}
\usepackage{amsmath}
\usepackage{amsthm}
\usepackage{booktabs}
\usepackage{algorithm}
\usepackage{algorithmic}
\usepackage[switch]{lineno}

\usepackage{verbatim}
\usepackage{svg}
\usepackage{enumitem}
\usepackage{lipsum}
\usepackage{amssymb}
\usepackage{booktabs}
\usepackage{multirow}
\usepackage{fixltx2e}
\usepackage{xcolor}
\usepackage{caption}

\usepackage{minitoc}

\usepackage{capt-of}
\usepackage{tocloft}
\usepackage{xpatch}

\newcommand{\listofappendixfiguresnameA}{\small{List of Figures}}
\newlistof{appendixfigures}{apf}{\listofappendixfiguresnameA}

\makeatletter
\newcommand\listofappendixfiguresA{%
  \chapter*{\listofappendixfiguresnameA
    \@mkboth{\MakeUppercase\listofappendixfiguresnameA}%
            {\MakeUppercase\listofappendixfiguresnameA}}%
  \@starttoc{apf}%
}
\makeatother

\newcommand{\listofappendixtablesnameA}{\small{List of Tables}}
\newlistof{appendixtables}{apt}{\listofappendixtablesnameA}
\makeatletter
\newcommand\listofappendixtablesA{%
  \chapter*{\listofappendixtablesnameA
    \@mkboth{\MakeUppercase\listofappendixtablesnameA}%
            {\MakeUppercase\listofappendixtablesnameA}}%
  \@starttoc{apf}%
}
\makeatother

\makeatletter
\xapptocmd{\appendix}{%
  \write\@auxout{%
    \string\let\string\latex@tf@lof\string\tf@lof
    \string\let\string\tf@lof\string\tf@apf%
    \string\let\string\latex@tf@lof\string\tf@lot
    \string\let\string\tf@lot\string\tf@apt%
  }%
}{}{}
\makeatother

\newcommand{\citet}[1]{\citeauthor{#1}, \citeyear{#1}}

\newcommand{\scriptworld}{\texttt{ScriptWorld}}

\title{ScriptWorld: Text Based Environment For Learning Procedural Knowledge}

\author{
Abhinav Joshi\and
Areeb Ahmad\and
Umang Pandey\and
Ashutosh Modi \\
\affiliations
Indian Institute of Technology Kanpur (IIT-K)
\emails
\{ajoshi, ashutoshm\}@cse.iitk.ac.in, \{areeb, umangp\}@iitk.ac.in
}

\begin{document}
\maketitle

\doparttoc 
\faketableofcontents 



\input{./abstract}
\input{./Introduction}
\input{./RelatedWorks}

\input{./ScriptWorldEnvironment}

\input{./ProposedBaselines}
\input{./Experiments}
\input{./Discussion}
\input{./Conclusion}

\bibliographystyle{named}
\bibliography{references}

\clearpage
\newpage

\appendix
\input{./Appendix.tex}

\end{document}

%% file: abstract.tex
\begin{abstract}
Text-based games provide a framework for developing natural language understanding and commonsense knowledge about the world in reinforcement learning based agents. Existing text-based environments often rely on fictional situations and characters to create a gaming framework and are far from real-world scenarios. In this paper, we introduce \scriptworld: a text-based environment for teaching agents about real-world daily chores and hence imparting commonsense knowledge. To the best of our knowledge, it is the first interactive text-based gaming framework that consists of daily real-world human activities designed using scripts dataset. We provide gaming environments for 10 daily activities and perform a detailed analysis of the proposed environment. We develop RL-based baseline models/agents to play the games in \scriptworld. To understand the role of language models in such environments, we leverage features obtained from pre-trained language models in the RL agents. Our experiments show that prior knowledge obtained from a pre-trained language model helps to solve real-world text-based gaming environments. 
\end{abstract}

%% file: Introduction.tex
\section{Introduction}\label{sec:intro}


\begin{figure}[t]
\centering
  \includegraphics[scale=0.28]{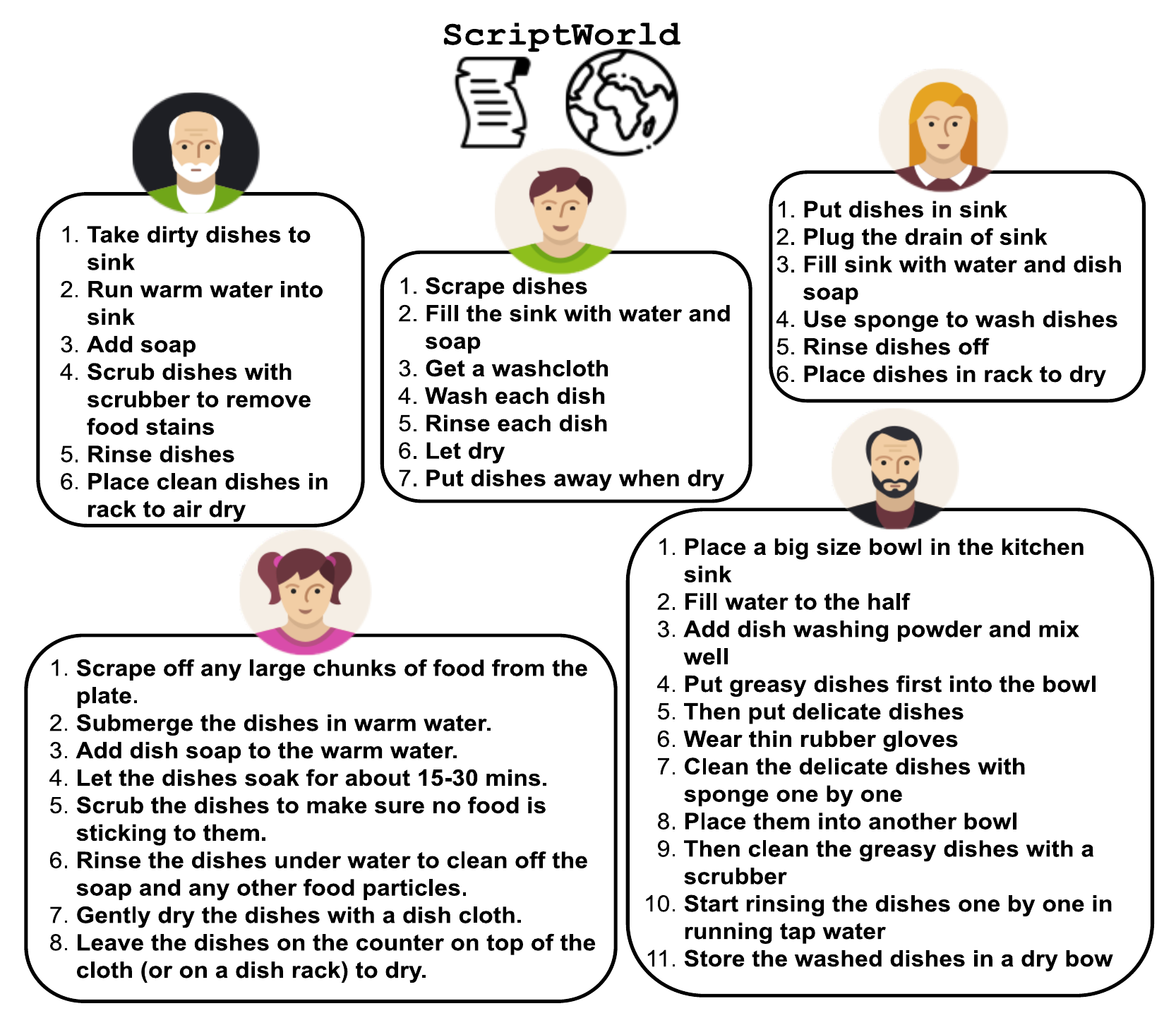}
    \caption{Different descriptions for the \texttt{Washing Dishes} script scenario.}
  \label{fig:script-example}
\end{figure}

Text-based games in reinforcement learning have attracted research interests in recent years \cite{hausknecht2020interactive,Kuttler2020TheNL}.
These games have been developed to impart Natural Language Understanding (NLU) and commonsense reasoning capabilities in Reinforcement Learning (RL) based agents. A typical text-based game consists of a textual description of states of an environment where the agent/player observes and understands the game state and context using text and interacts with the environment using textual commands (actions). For successfully solving a text-based game, in addition to language understanding, an agent needs complex decision-making abilities, memory, planning, questioning, and commonsense knowledge \cite{cote2018textworld}. Existing text-based gaming frameworks (e.g., Jericho \cite{hausknecht2020interactive}) provide a rich fictional setup (e.g., treasure hunt in a fantasy world) and require an agent to take complex decisions involving language and fantasy world knowledge. However, the existing text-based frameworks are created using a fixed prototype and are often distant from real-world scenarios involving daily human activities.  Though these frameworks aim to provide a rich training bench for enhancing NLU in RL algorithms, the fictional concepts in these games are not well grounded in real-world scenarios, making the learned knowledge non-applicable to the real world. In contrast, for trained RL algorithms to be of practical utility, they should be trained in real-world scenarios that involve daily human activities. Humans carry out daily activities (e.g., making coffee, going for a bath) without much effort by making use of implicit \textit{Script Knowledge}. 

\noindent Formally, \textbf{Scripts} are defined as sequences of actions describing stereotypical human activities, for example, cooking pasta, making coffee, etc. \cite{schank1975scripts}. Scripts entail knowledge about the world. For example, when someone talks about ``Washing Dishes", there lies an implicit knowledge of fine-grained steps which would be present in the activity. By just saying, “I washed dishes on Thursday,” a person conveys the implicit knowledge about the entire process (Fig. \ref{fig:script-example}).  
The detailed implicit understanding of a task not only helps to learn about an activity but also facilitates taking suitable actions depending on the environment and past choices. Moreover, for learning a new task, humans can quickly and effortlessly discover new skills for performing the task either by their knowledge about the world or by reading (a manual) about it. With the aim to promote similar learning behavior in RL agents, in this paper, we propose \textbf{\scriptworld}, a new text-based game environment based on real-world scenarios involving script knowledge. 

\noindent The motivation for creating \scriptworld\ environment is threefold. Firstly, \scriptworld\ environment is based on the concept of scripts that encapsulates commonsense and procedural knowledge about the world. The environment is designed to enable agents to learn this knowledge while participating in the game. Scripts have non-linear structure \cite{wanzare-etal-2016-crowdsourced}. A script scenario can be described in multiple ways with linguistic variation across different descriptions. Fig. \ref{fig:script-example} shows different descriptions for the washing dishes scenario. Moreover, at the level of execution, the order of events/actions within the script can vary across different descriptions of a scenario. For example, some events may be skipped, and the order of events might vary. Hence, learning script knowledge is challenging. Taking into account the variability in descriptions of a scenario, an agent needs to learn the prototypical order of events and needs to abstract out the meaning of different verbal descriptions of an action. Secondly, \scriptworld\ being a text-based environment about everyday scenarios, provides an opportunity for grounded language learning and understanding. Language phenomena do not happen in isolation, but the semantics are grounded in the real world \cite{groundedLanguage2017}; \scriptworld\ provides the environment to establish and learn that grounding. Lastly, there have been extensive studies that have explored the cognitive basis of script knowledge in humans \cite{miikkulainen1993subsymbolic,Modi17-Thesis}. 
\scriptworld\ involves the acquisition of script knowledge. Consequently, it provides an opportunity to compare the behavior of a trained RL agent with humans providing further insights into the cognitive aspects. 

\noindent In a nutshell, we make the following contributions: 
\begin{itemize}[noitemsep,topsep=0pt]
\item We introduce a new interactive text-based gaming environment, \scriptworld consisting of games based on script descriptions provided by human annotators for performing realistic daily chores. We perform a detailed analysis of the proposed environment and release the environment and agents: \url{https://github.com/Exploration-Lab/ScriptWorld}. 
\item We propose and experiment with a battery of  Reinforcement Learning (RL) agents based on pre-trained Language Models (LM) as baselines for solving the \scriptworld\ environment. The experiments show that pre-trained LMs, when combined with RL agents, give reasonable performance, pointing towards scope for improvement and inclusion of prior knowledge. 
\end{itemize}

%% file: RelatedWorks.tex
\section{Related Work} \label{sec:relatedwork}

\paragraph{Text Based Games.} 
Text-based games are divided into three main categories based on how an agent/player might issue (take) commands (actions): Parser-based, Choice Base, and Hyper Text Based \cite{he-etal-2016-deep}. The player issues a command in Parser-based games by typing in the input, and an inbuilt parser parses it. In Hypertext-based games, the player issues a command by selecting one of the Hyperlinks present in the prompt. In choice-based games, the player chooses the command from a list of options in addition to the state description. Parser-based games are limited since these can only parse sentences that adhere to pre-defined grammar and vocabulary. Giving flexibility for free-form text suffers from the exponentially increasing action space. \scriptworld\ uses choice-based approach (also see \S \ref{sec:discussion}). Moreover, in general, choice-based games are more popular among humans than parser-based games \cite{he-etal-2016-deep}. \citet{cote2018textworld} have introduced TextWorld sandbox environment, a Python-based framework in which the user can build parser-based game worlds of varying difficulty along with in-game objects and goal states while monitoring states and assigning rewards. Language diversity and complexity of action space are limited in TextWorld. In contrast, \scriptworld\ (created using human written texts) overcomes these issues by generating ample alternative pathways to complete a task. The complexity and variability in \scriptworld\ help to develop better language understanding capabilities in agents. 
Other Text-based game frameworks have been proposed, such as TWC (TextWorld Commonsense) \cite{Murugesan2021TextbasedRA}, and Question Answering with Interactive Text (QAit) \cite{Yuan2019InteractiveLL} build on TextWorld. Similarly, \citet{hausknecht2020interactive} have introduced a new framework called Jericho, which facilitates using man-made Interactive Fiction Games as learning environments for agents to train and learn. 

\paragraph{Scripts.}
Scripts have been an active area of research for the last four decades. As evident from the definition (\S\ref{sec:intro}), scripts encapsulate commonsense and procedural knowledge about the world and hence are an ideal source for training agents to learn about the world. Several computational models have developed for modeling script knowledge, inter alia, \cite{regneri2010learning,frermann2014hierarchical,modi-2016-event,modi-titov-2014-inducing,rudinger-etal-2015-learning,jans-etal-2012-skip,pichotta-mooney-2016-using,modi-etal-2017-modeling}. A number of corpora have also been created, e.g., InScript \cite{modi-etal-2016-inscript}, DeScript \cite{wanzare-etal-2016-crowdsourced}, McScript \cite{ostermann-etal-2018-mcscript,ostermann-etal-2018-semeval}, and ProScript \cite{sakaguchi-etal-2021-proscript-partially}. Researchers have also examined script knowledge from the perspective of language modeling \cite{lm-scripts-2021}. 

\paragraph{RL Agents.}
\citet{narasimhan2015language} have introduced an RL-based architecture called LSTM-DQN that learns the action policies and state representations of parser-based games. A number of other agents have been proposed for text-based environments, e.g., \citet{he-etal-2016-deep} have introduced DRRN (Deep Reinforcement Relevance Network) architecture, KG-DQN architecture \cite{ammanabrolu-riedl-2019-transfer,ammanabrolu2020graph,10.5555/3495724.3495980,chaudhury2020crest,Adolphs2020LeDeepChefDR,Yin2019LearnHT,Yao2020KeepCA}. \citet{Singh2022PretrainedLM} introduce a pretrained language model finetuned on the dynamics of the game to equip the agent with language learning capabilities as well as acquire real-world knowledge. Our baseline agents come close to \citet{Singh2022PretrainedLM}.


\begin{table}[t]
\centering
\tiny
\renewcommand{\arraystretch}{1}
\setlength\tabcolsep{2pt}
\begin{tabular}{cccc}
\toprule
Scenario                          & Nodes & \begin{tabular}[c]{@{}c@{}} Deg. \end{tabular} & Paths               \\
\midrule
\texttt{Taking a \textbf{Bath}}                     & 525   & 3.7                                                         & $3.1e+{27}$ \\
\texttt{Baking a \textbf{Cake}}                     & 542   & 3.6                                                         & $4.0e+{26}$ \\
\texttt{Flying in an \textbf{Airplane}}            & 528   & 3.6                                                         & $2.6e+{30}$ \\
\texttt{Going Grocery \textbf{Shopping}}   & 544   & 3.7                                                         & $2.3e+{26}$ \\
\texttt{Going on a \textbf{Train}}                   & 427   & 3.7                                                         & $3.1e+{21}$ \\
\texttt{Planting a \textbf{Tree}}                   & 373   & 3.7                                                         & $1.6e+{16}$ \\
\texttt{Riding on a \textbf{Bus}}                  & 376   & 3.8                                                         & $1.0e+{17}$ \\
\texttt{Repairing Flat \textbf{Bicycle} Tire}     & 402   & 3.4                                                         & $8.4e+{18}$ \\
\texttt{Borrowing Book from \textbf{Library}} & 397   & 3.7                                                         & $3.1e+{19}$ \\
\texttt{Getting a \textbf{Haircut}}                & 528   & 3.7                                                         & $4.0e+{28}$\\
\bottomrule
\end{tabular}
\caption{The table compares graphs of different scenarios present in \scriptworld. Deg. represents the average degree for the nodes in the scenario graph.}
\label{tab:scenario_graph_comparison}
\end{table}


\begin{figure*}[t]
\centering
  \includegraphics[scale=0.28]{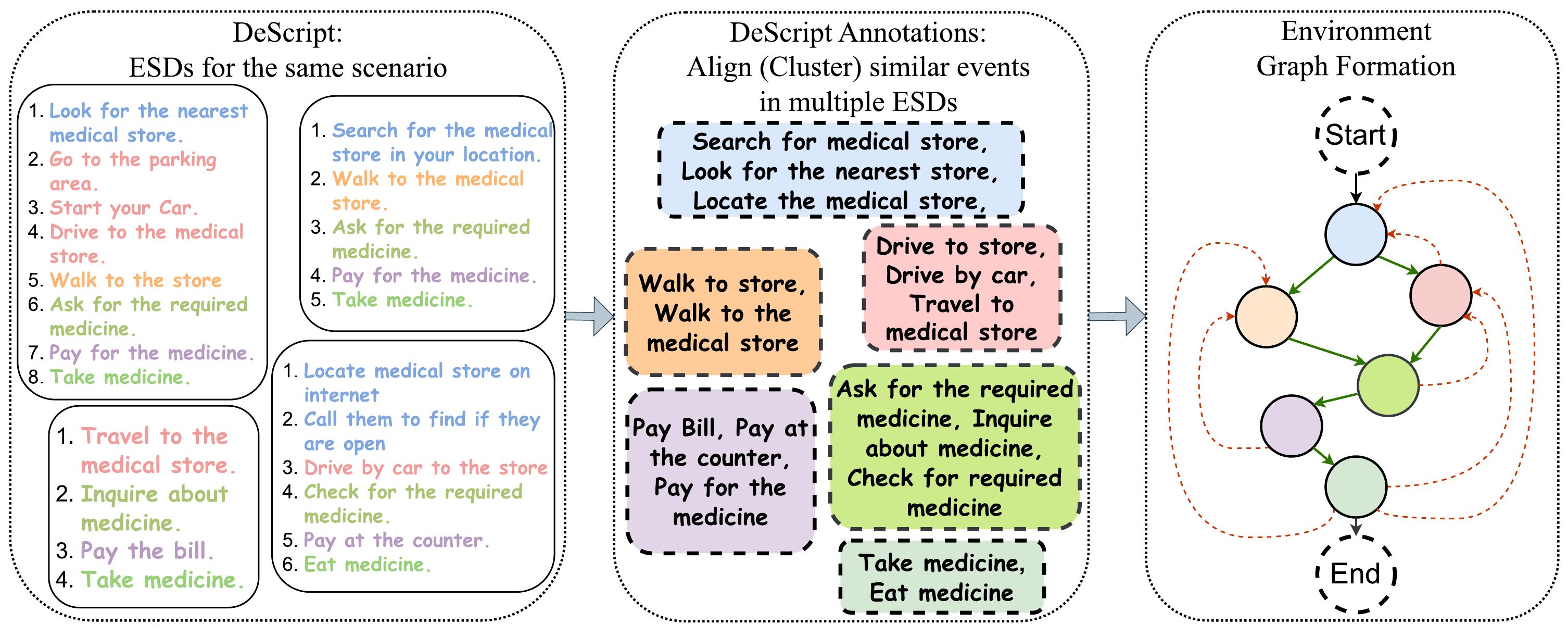}
  \caption{The figure shows a simplified version of the  scenario, \texttt{Get Medicine}, and the process of creating an environment graph (right diag.) from the ESDs (left diag.) and aligned events (middle diag.) for the scenario. The green directed edges in the environment graph represent the correct paths, and the red edges denote the environment transition when a wrong option is selected.}
  \label{fig:script-graph}
\end{figure*}

\begin{figure}[t]
\centering
  \includegraphics[scale=0.50]{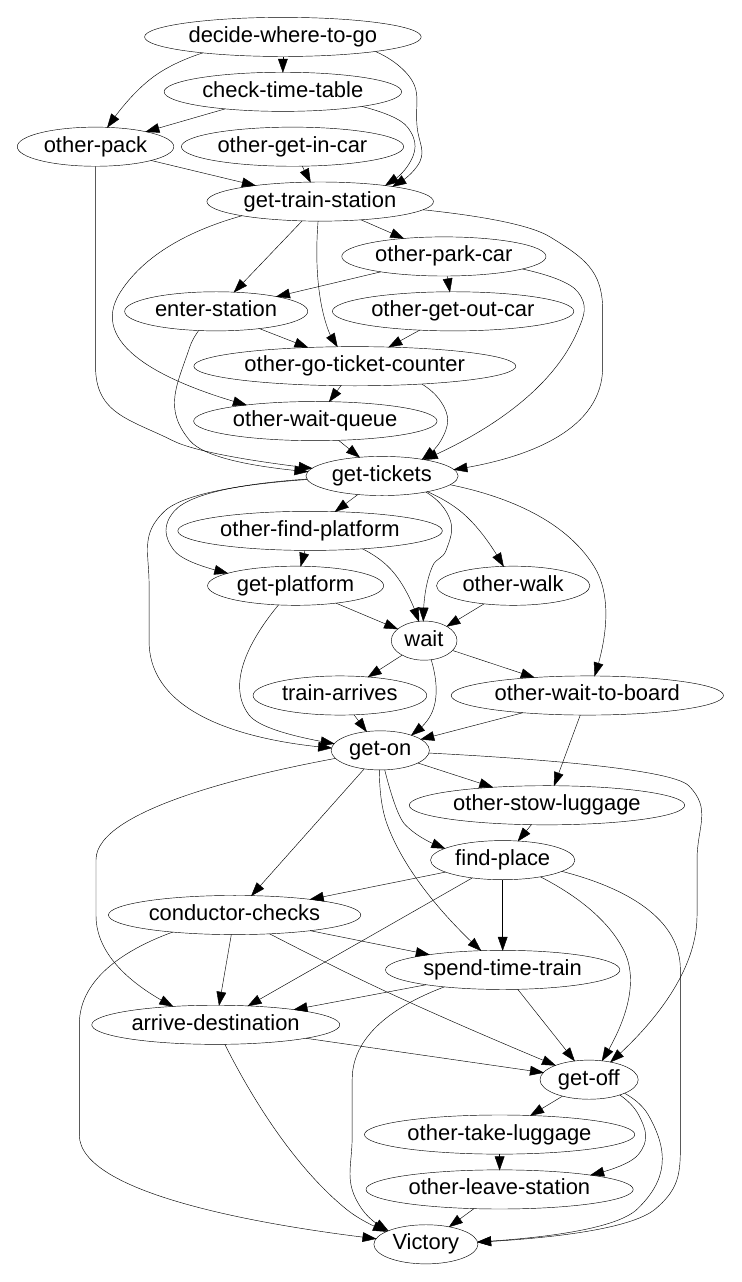}
  \caption{{The figure shows the \textit{``compact graph"} created for the scenario \texttt{Going on a Train}. }}
  \label{fig:train_compact_graph_main}
\end{figure}

%% file: ScriptWorldEnvironment.tex
\section{\scriptworld\ Environment} \label{sec:scr-env}

\scriptworld\ tries to bridge the gap between real-world scenarios (via Scripts) and text-based games for RL by creating a suitable environment. We take into consideration three design choices for developing the environment: \textbf{1) Complexity:} The game environment should be complex enough to test an RL algorithm’s capacity to capture,  understand and remember reasonable steps required for performing a daily chore. \textbf{2) Flexibility:} For an environment to help develop and debug RL algorithms, it becomes imperative to consider flexibility as a feature. The environment should be flexible regarding difficulty levels and handicaps (hints) to provide a good test bench for reinforcement learning algorithms. \textbf{3) Relation to Real-World scenarios:} The environment should consist of activities/tasks grounded in the real world and well understood among humans.

\paragraph{DeScript.} 
Given the nature of Script knowledge, we use a scripts corpus referred to as DeScript \cite{wanzare-etal-2016-crowdsourced} for creating \scriptworld\ environment. DeScript is a corpus having a telegram-style sequential description of a scenario in English (e.g., baking a cake, taking a bath, etc.) DeScript is created via crowd-sourcing. For a given scenario, crowd-workers write a point-wise and sequential short description of various events involved in executing the scenario (this one complete description is called an ESD (Event Sequence Description)). Fig \ref{fig:script-example} shows an example of 5 ESDs for the \texttt{Washing Dishes} scenario. DeScript collects data for 40 daily activities (scenarios), and 100 ESDs (written by different crowd-sourced workers) are collected for each scenario. Additionally, for a given scenario, semantically similar events from different ESDs are manually aligned by human annotators (more details about data collection and annotations are present in \citet{wanzare-etal-2016-crowdsourced}). The alignment annotation is done for 10 scenarios (Table \ref{tab:scenario_graph_comparison} gives the list of scenarios). In the present version of \scriptworld, we only include these 10 scenarios with gold alignment. Another line of work can be to consider sequence alignment algorithms \cite{chatzou2016multiple} to align sequences for the remaining 30 scenarios. However, as observed in initial experiments, the error rate of alignment algorithms gets propagated to the graph formation leading to a less reliable environment. We leave the automatic alignment of the remaining 30 scenarios for future work. The gold alignments in the DeScsript corpus contain cluster annotations of similar events across multiple ESDs into a single abstract, generalized event. For example, Fig. \ref{fig:script-graph} depicts the scenario, \texttt{Get Medicine}, where similar events from ESDs written by different people are clustered to form generalized event categories. Further, the combined set of events and the relation between the ESDs is leveraged to construct a graph (as explained later) where each node represents an abstract event. To the best of our knowledge, the proposed method is the first novel approach to create an environment (based on script knowledge) that could be useful for training RL agents. 

\noindent The \scriptworld\ environment is created from scratch using Python. A typical game begins by providing a quest (goal) to the agent. The quest/goal is a one-line description of the scenario (e.g., plant a tree). The agent is also provided with initial observations (in English). Since it is a choice-based game, at each step in the game, the agent is also presented with a list of actions/choices (in English) that it could opt to advance towards the goal. Based on the action selected by the agent, it is awarded a zero/positive/negative reward at each step. Every correct action takes the agent closer to task completion, whereas every wrong action results in a deviated path. 
(also see App. A, 
the appendix is available at \url{https://github.com/Exploration-Lab/ScriptWorld}). 



\paragraph{Graph Formation.} DeScript provides set of aligned ESDs  ($\mathcal{E}_{1}^{\mathcal{S}_i}, \mathcal{E}_{2}^{\mathcal{S}_i},\ldots,\mathcal{E}_{N}^{\mathcal{S}_i}$) for a scenario $\mathcal{S}_i$. Each ESD $\mathcal{E}_{k}^{i}$ consists of sequence of short event descriptions: $\mathbf{e}_1^{(\mathcal{E}_{k}^{i})}, \mathbf{e}_2^{(\mathcal{E}_{k}^{i})}, \ldots \mathbf{e}_n^{(\mathcal{E}_{k}^{i})}$. Gold alignment in DeScript results in events in different ESDs that are semantically similar, getting linked to each other, i.e., clustered together. For example, for the \texttt{Washing Dishes} scenario, events ``put dishes in sink'' ($\mathbf{e}_1^{(\mathcal{E}_{1}^{Wash})}$) in $\mathcal{E}_{1}^{Wash}$ and ``take dirty dishes to sink'' ($\mathbf{e}_1^{(\mathcal{E}_{2}^{Wash})}$) in $\mathcal{E}_{2}^{Wash}$ are linked (clustered) together. Aligned events (from different ESDs) are used to create a graph having nodes as the event clusters (of aligned events) and directed edges representing the prototypical order of the events. In particular, a directed edge is drawn from node $p$ to $q$ if there is at least one event in node $p$ that directly precedes an event in node $q$. We refer to the created event node graph as the \textit{compact graph} (Fig. \ref{fig:train_compact_graph_main}), compact graphs for other scenarios are in App. A. 
The alignment annotations in the DeScript also group multiple sets of actions that belong to the same event. For example, an event ``go to the terrace'' can be performed in two sets of sequenced steps by different annotators. 1) call the elevator $\rightarrow$ step in elevator $\rightarrow$ step out at the top floor, and 2) find stairs $\rightarrow$ climb stairs $\rightarrow$ reach top floor. We leverage the presence of such instances in the graph node to enrich the complexity of our environment. We split each event node in the compact graph into two nodes, the entry event node and the exit event node. Further, multiple action sequences result in parallel paths for reaching the exit node from the entry node (see also App. A). 
For instance, the above example will result in two parallel paths, where a player or an agent has to decide at the entry node to either take the elevator or the stairs. If players choose to take the stairs, they are expected to follow the next set of actions to reach the terrace. Moreover, all the sub-steps in this event now result in multiple graph nodes. We refer to this graph as the \textit{scenario graph} (see App. A). 
This helps to capture the variability in performing daily chores, making the environment more realistic. Though the DeScript corpus provides clustered events for every scenario, after graph creation, we found that a few of the ESDs present in the corpus were inconsistent, not fitting the commonsense reasoning for a procedure. We also observed that some of the ESDs written by annotators are too small and describe the task in generic terms. Such ESDs, when considered in graph formation, result in direct paths to the final goal node, making the game less complex. We remove all such inconsistencies from the graph by manual inspection, making it more reliable for capturing script knowledge and keeping the realism intact for the environment. The compact graph serves as an initial starting point for creating the scenario graph. The agents are trained on a scenario graph. 

\noindent To quantitatively capture the complexity of scenarios in \scriptworld, we calculate the total number of paths reaching the end node from the start node. We first compute the total number of paths in the compact graph using a depth-first traversal. Further, we extend the computation by adding the number of parallel paths present for each entry and exit event node in the scenario graph. $\text{TotalPaths} = \sum\limits_{p_k=0}^T \prod\limits_{i=1}^{N} {n_i}^{(p_k)}$, where $T$ is the total number of paths in a compact graph, $N$ represents the total number of nodes in a path $p_k$ and ${n_i}^{(p_k)}$ denotes the number of splits for the $i^{\text{th}}$ node. Table \ref{tab:scenario_graph_comparison} shows the total number of paths. As evident from the table, the number of paths in each of the scenarios is enormous and demonstrates the highly complex nature of the environment. Overall, the scenario \texttt{Flying in an Airplane} turns out to be the most complex one in terms of the number of correct possible paths. This is possibly due to more variability in carrying out this activity. 

\paragraph{Environment Creation.} We create the game environment using scenario graphs. For each state in the environment, the agent is required to pick the correct action (choice) from the available options. Since the created scenario graph contains a wide variety of suitable actions grouped in a node, we sample the right choice from the available actions in a node. Note that sampling of correct actions happens randomly at every visit, making the environment highly dynamic. To create incorrect choices, we exploit the temporal nature of the scenario graphs. As a scenario graph contains the sequence of actions to perform a specific sub-task, all actions in nodes (both past as well as future nodes are considered) that are far from the current node become invalid for the current state. For selecting this node distance, we manually experiment with different node distances and find the different distances ($d_1, d_2, \ldots d_{10}$) suitable for sampling the invalid actions, i.e., for a scenario $i$, we consider all nodes at a distance greater than $d_i$ hops from the current node (Table in App. A 
shows various distances chosen for each of the scenarios). This strategy of sampling the invalid choices makes the environment more complex as all the options are related to the same scenario, and an understanding of event order in a task is required to achieve the goal. 

\noindent\textbf{Rewards (Performance Scores):} For all the scenarios, every incorrect action choice results in a negative reward of -1, and every correct choice returns a 0 reward. For task completion, the agent gets a reward of 10, i.e., a player gets a maximum reward of 10 at the end of each game if they choose a correct sequence of actions. The choice of zero rewards for correct action helps RL algorithms explore multiple correct ways of performing a task, capturing the generalized procedural knowledge required for a specific task. The game terminates when an agent chooses 5 successive wrong actions.  

\begin{table*}[t]
\centering
\tiny
\renewcommand{\arraystretch}{0.9}
\setlength\tabcolsep{5pt}
\begin{tabular}{ccccccccc}
\toprule
\multirow{2}{*}{Algorithm}        & \multicolumn{2}{c}{DQN} & \multicolumn{2}{c}{A2C} & \multicolumn{2}{c}{PPO} & \multicolumn{2}{c}{RPPO} \\
\cmidrule{2-9}
                                  & handicap      & w/o handicap    & handicap      & w/o handicap     & handicap          & w/o  handicap         & handicap      & w/o  handicap     \\
                                  \midrule
\texttt{Shopping}            & 9.60 ( $\pm$ 0.62)       & -7.28 ($\pm$ 13.15)      & 
9.90 ($\pm$ 0.30)      & -9.81 ($\pm$ 14.71)       & 9.84 ($\pm$ 0.39)         &-4.78 ($\pm$ 10.79)          &  9.71 ($\pm $ 0.57)       & \textbf{8.79 ($\pm $ 4.15)}       \\
\texttt{Bus}                   & 8.98 ($\pm$ 0.79)      & -1.47 ($\pm$ 11.16)      & 9.89 ($\pm$ 0.34)      & -7.37 ($\pm$ 17.09)       & 9.93 ($\pm$ 0.25)          & 1.50 ($\pm$ 7.50)           & 9.97 ($\pm $ 0.17)      & \textbf{9.32 ($\pm $ 1.24)}       \\
\texttt{Train}                 & 9.21 ($\pm$ 2.07)       & -3.10 ($\pm$ 11.16)      & 9.89 ($\pm$ 0.31)       & -8.13 ($\pm$ 14.99)       & 9.75 ($\pm$ 0.49)           & -1.13 ($\pm$ 9.47)           & 9.56 ($\pm $ 0.80)      & \textbf{8.19 ($\pm $ 4.70)}        \\
\texttt{Library} & 9.51 ($\pm$ 0.68)       & -1.94 ($\pm$ 9.87)      & 9.88 ($\pm$ 0.32)      & -3.03 ($\pm$ 9.84)      & 9.90 ($\pm$ 0.30)          & 1.12 ($\pm$ 7.31)           & 9.89 ($\pm $ 0.31)      & \textbf{8.41 ($\pm $ 4.77)}       \\
\texttt{Haircut}                & 9.88 ($\pm$ 0.35)      & -9.30 ($\pm$ 12.93)      & 9.89 ($\pm$ 0.34)      & -5.87 ($\pm$ 12.28)       & 9.85 ($\pm$ 0.38)         & -4.30 ($\pm$ 10.84)           &  9.63 ($\pm $ 0.64)      & \textbf{6.32 ($\pm $ 5.29)}      \\

\texttt{Cake}                     & 9.32 ($\pm$ 0.84)      & -4.13 ($\pm$ 9.22)      & 9.48 ($\pm$ 0.92)     &-7.58 ($\pm$ 13.18)       & 9.87 ($\pm$ 0.34)         & -4.46 ($\pm$ 12.32)            & 9.78 ($\pm $ 0.48)     & \textbf{7.18 ($\pm $ 4.97)}       \\
\texttt{Bicycle}     & 9.50 ($\pm$ 0.75)       & 0.07 ($\pm$ 7.89)      & 9.95 ($\pm$ 0.22)       & -3.49 ($\pm$ 12.39)        & 9.90 ($\pm$ 0.33)           & 1.17 ($\pm$ 6.93)          &  9.74 ($\pm $ 0.57)      & \textbf{7.85 ($\pm $ 5.12)}       \\
\texttt{Tree}                   & 9.94 ($\pm$ 0.24)       & -0.15 ($\pm$ 7.83)   & 9.86 ($\pm$ 0.44)       & -3.54 ($\pm$ 12.56)       & 9.98 ($\pm$ 0.14)           & 1.43 ($\pm$ 7.29)           & 9.96 ($\pm $ 0.19)      & \textbf{8.88 ($\pm $ 3.23)}       \\

\texttt{Airplane}             & 9.68 ($\pm$ 0.75)      & -4.21 ($\pm$ 12.39)      & 9.86 ($\pm$ 0.35)      & -8.66 ($\pm$  12.66)        & 9.86 ($\pm$ 0.40)           & -4.74 ($\pm$ 11.08)          & 9.54 ($\pm $ 0.73)    & \textbf{6.85 ($\pm $ 6.12)}       \\

\texttt{Bath}                     & 9.68 ($\pm$ 0.61)      & -6.49 ($\pm$ 13.23)     &9.75 ($\pm$ 0.57)& -10.02 ($\pm$ 15.95)       & 9.84 ($\pm$ 0.37)         & -5.35  ($\pm$ 11.19)           & 9.45 ($\pm $ 0.82)     & \textbf{6.35 ($\pm $ 5.59)}      \\
\bottomrule
\end{tabular}
\caption{The table shows performance scores (averaged over multiple runs) of various agents for all the scenarios (number of choices = 2). The number in brackets shows the standard deviation of the score. Paraphrase Albert Small V2 is used as the LM}
\label{tab:agent_results_on_all_scenarios}
\end{table*}

\noindent\textbf{Flexibility: }To introduce flexibility in \scriptworld, we consider two settings in a game. \textbf{1) }\textbf{Number of choices:} At each step, the number of choices presented to an agent can be changed (1 correct choice and the rest all incorrect). As the number of options increases, it becomes more challenging for an agent to choose the right action. \textbf{ 2)} \textbf{Number of backward hops for wrong actions:} We choose the number of backward hops as another game setting that decides how many hops to displace whenever a wrong action is selected. When an agent selects an incorrect choice, its location is displaced by hopping it backward in the temporal domain, and this back-hop distance is another parameter in the environment. In our experiments, agents played with the environment with a back-hop distance of 1. Due to the presence of parallel paths in the graph, an agent hops to a previous node in case of incorrect action and may not follow the same path again, which acts as a penalty. For the start node, since backward hop is not possible, the agent remains at the same position; however, both positive and negative choices are re-sampled, and consequently, observations change. These parameters introduce flexibility in our environment, giving the freedom to create a suitable test bench for RL algorithms.

\noindent\textbf{Handicaps (Hints):} Text-based games are often challenging for RL agents playing from scratch. To mitigate the complexity issue, we introduce a version of the game with hints (referred to as handicaps) for each state. The hint for a state provides a short textual clue for the next action to take at the current state. The presence of hints in the environment makes the gameplay relatively easier. Hints are generated automatically using GPT2 \cite{radford2019language}. Scenario title concatenated with state node event description (separated by a full-stop) is given as the prompt to GPT2 for generating a large number of hints, and then a hint is sampled from them. We manually examined the hints to ensure they did not repeat (verbatim) any of the existing actions. To introduce variability, one could also stochastically decide to show a hint, e.g., by sampling from a Bernoulli distribution at each state. However, in this paper, we consider only the setting where hints are shown at every state. We leave this for future work.

\noindent\textbf{Comparison with other text-based environments: } \scriptworld\ environment is different from the existing text-world-based environments (e.g., Text World, Jericho, TWC, QAit). The primary novelty of \scriptworld\ comes from the inclusion of realistic scenarios made by leveraging ESDs written by human annotators, and this requires procedural knowledge to solve the game. The complexity (Table \ref{tab:scenario_graph_comparison}) of the \scriptworld\  is much more than the existing environments, requiring the agent to remember past events and actions. We provide more details about \scriptworld\ and compare it with other environments in App. A. 

%% file: ProposedBaselines.tex
\section{RL Baselines}\label{sec:baselines}
In the \scriptworld\ environment, for every state, the environment returns a sample of a possible set of choices. Since these choices provide feedback related to the current state, the agent must keep track of all the observations received after a particular choice. This property typically resembles the Partially Observable Markov decision processes (POMDP) \cite{pomdp}, where the agent can never observe the complete state of the environment. Formally, \scriptworld\ is defined by $(S, A, \Omega, R, \gamma)$, where $S$ is the set of environment states (nodes in the scenario graph), and $A$ is the set of all actions (choices), $\Omega$ is the set of observations, i.e., description of various actions, $R$ is the reward obtained and $\gamma$ is the discount parameter. The goal of an agent is to learn a policy $\pi(a \mid s)$, i.e., a mapping from a set of observations to actions leading to an optimal choice in a particular state. In some algorithms (e.g., DQN: Deep Q-Network), instead of learning the policy, the agent learns q-values, which can reveal the policy. Formally, q-value (q-function) $Q_{\pi}(s,a)$ is the expected cumulative return if an agent starts from state $s$ and takes action $a$ and thereafter follows a policy $\pi$. Recent developments in RL have proposed an approximation of $\pi(a \mid s)$/q-value via a parameterized model that takes state (features) and actions (features) as input and produces the $\pi(a \mid s)$/q-value as the output \cite{sutton2018reinforcement}. We follow the same approach. 

\noindent Recently, Language Models (LM) have shown promising results in almost all tasks in NLP (e.g., \cite{lm-scripts-2021}). For the RL baselines for the \scriptworld\ environment, we consider using pre-trained SBERT language models \cite{reimers-2019-sentence-bert-sbert} as a source of prior real-world knowledge, which could be used directly by an RL algorithm to solve the environment. We consider a generalized scheme where a pre-trained language model extracts information from observations, i.e., the features extracted ($h_i = \text{LM}(c_i)$) from the available set of choices $c \in \{c_1,\ldots, c_n\}$) is used by the RL algorithms as input features. The pre-trained language model generates embeddings ($h_{i}$) corresponding to each of the provided $n$ options. The obtained embeddings are concatenated ($O$) and passed as input to the RL algorithm, i.e., \ \ \ \ \ \ \ \ \ \ \ \ \ \ \  $c \in \{c_1,\ldots, c_n\}; \ \ h_i = \text{LM}(c_i)$ 
$$O = h_1 \oplus h_2 \oplus \ldots \oplus h_n$$

\begin{figure}[t]
\centering
  \includegraphics[scale=0.18]{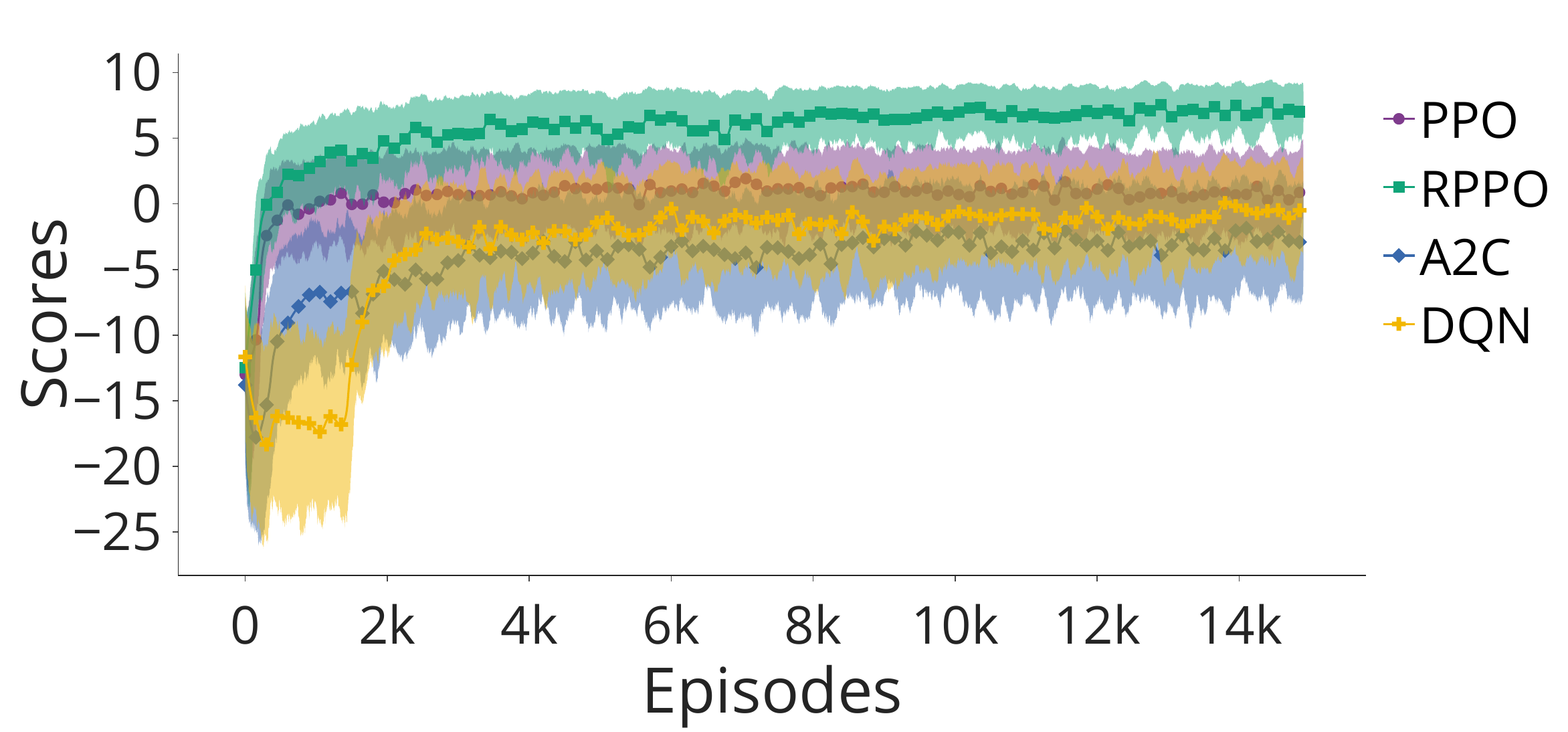}
  \caption{{The figure shows the performance comparison of multiple RL algorithms on scenario \texttt{Repairing a Flat Bicycle Tire} on setting (without handicap, choices=2).} Paraphrase Albert Small V2 is used as the LM. The plot shows moving average of performance curves across various episodes}
  \label{fig:All_agent_2woh_bicycle_para_albert}
\end{figure}

\noindent Subsequently, the RL framework generates $\pi(a \mid s)$/Q values for the available set of actions. With the help of this generalized architecture, we run a detailed set of experiments with combinations of multiple language models and different RL algorithms. In particular, we use DQN \cite{mnih2013playing}, A2C \cite{mnih2016asynchronous_A2C_A3C}, PPO \cite{PPO}, and RPPO: Recurrent PPO (PPO + LSTM). More details about RL agents, training, and other settings are provided in App. B. 
Some of the other existing works for language-based RL algorithms use knowledge-based agents. As these KBs do not directly adapt to our setting, we could not experiment with these approaches. In the future, we would explore how to make use of external knowledge to incorporate into the agent. 

\begin{figure}[t]
\centering
  \includegraphics[scale=0.18]{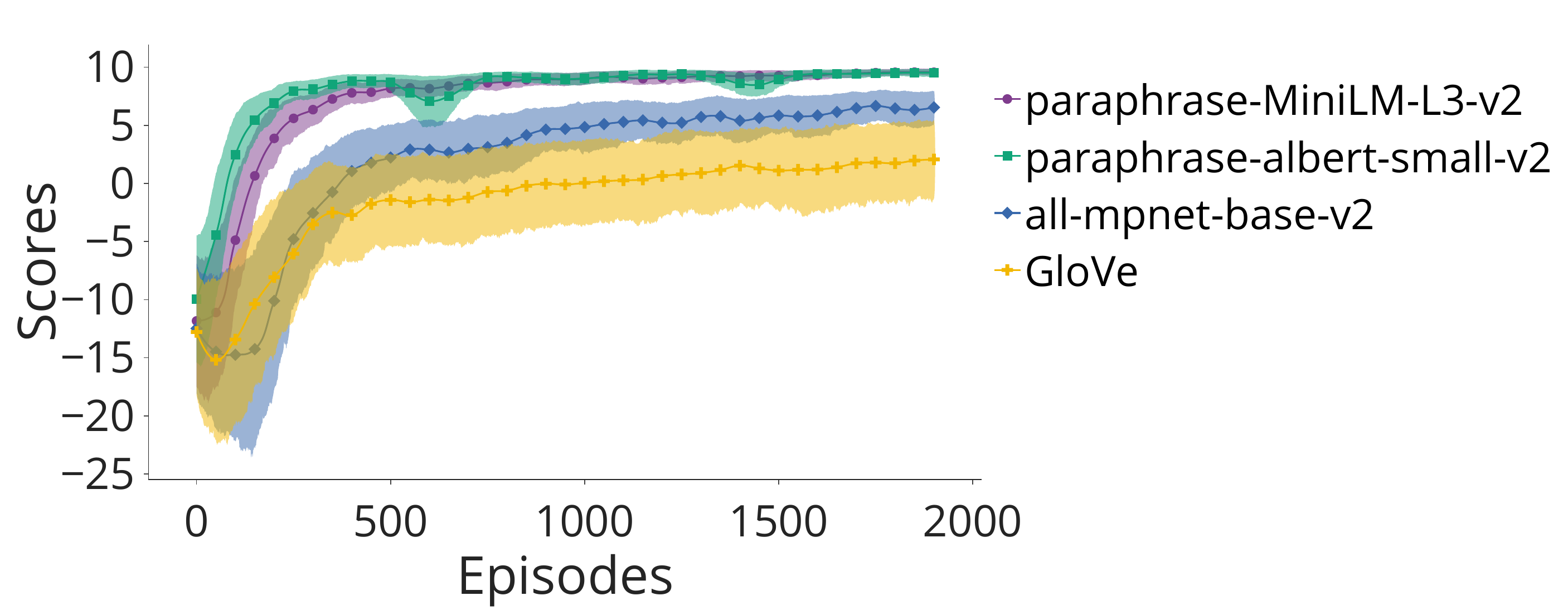}
  \caption{The figure shows the performance of the RPPO algorithm with various language models on scenario \texttt{Repairing a Flat Bicycle Tire} on setting (with handicap, choices=2).
  (highlighting the importance of LMs (contextual embeddings) over GloVe (non-contextual)).}
  \label{fig:LM_RPPO_2wh_bicycle}
\end{figure}

%% file: Experiments.tex
\section{Experiments, Results and Analysis} \label{sec:experiments}

\noindent\textbf{RL Agents Performances:} To benchmark the performance of existing RL algorithms on \scriptworld\, we perform an extensive set of experiments considering various combinations of language model embeddings and popular RL algorithms. Due to space limitations, we report the primary findings here, and the remaining are discussed in the App. D. 
Table \ref{tab:agent_results_on_all_scenarios} shows the performance of various RL algorithms in all the scenarios. The performance score is the score (total reward) achieved by an agent till the point of termination. As \scriptworld\ was designed, keeping flexibility the primary feature, in Table \ref{tab:agent_results_on_all_scenarios}, we report the performance of RL algorithms using multiple flexibility settings, i.e., with/without handicap and action choices = 2. The performance of algorithms with a handicapped version of the environment seems to be easier when compared to a non-handicapped version, depicting the choice of keeping the handicap feature to be useful. For settings without any handicap provided, we found the RPPO algorithm to beat other RL algorithms by a significant margin. Fig.  \ref{fig:All_agent_2woh_bicycle_para_albert} shows the performance of algorithms over multiple episodes, depicting the convergence rate. We observe that RPPO convergence is faster at a higher score, and DQN seems unstable during initial episodes. We also plot performance curves for all the scenarios in App. D. 
As our RL framework combines language embeddings with RL algorithms, we also highlight the effect of different language model embeddings. We choose RPPO for reporting performance with different language models, as in extensive experimentation, we found RPPO to perform better than other RL algorithms on multiple environment settings. Fig. \ref{fig:LM_RPPO_2wh_bicycle} reports the RPPO performance with different embeddings. We consider various types of SBERT-based embeddings (\url{https://www.sbert.net/docs/pretrained_models.html}). To judge the effect of contextual embeddings, we also report the RPPO performance with GloVe embeddings \cite{pennington2014glove}. RPPO with GloVe embeddings (non-contextualized word representations) performs poorly, depicting the importance of the context which is captured by contextualized LMs (more results on LMs in App. D.) 

\begin{figure}[t]
\centering
  \includegraphics[scale=0.20]{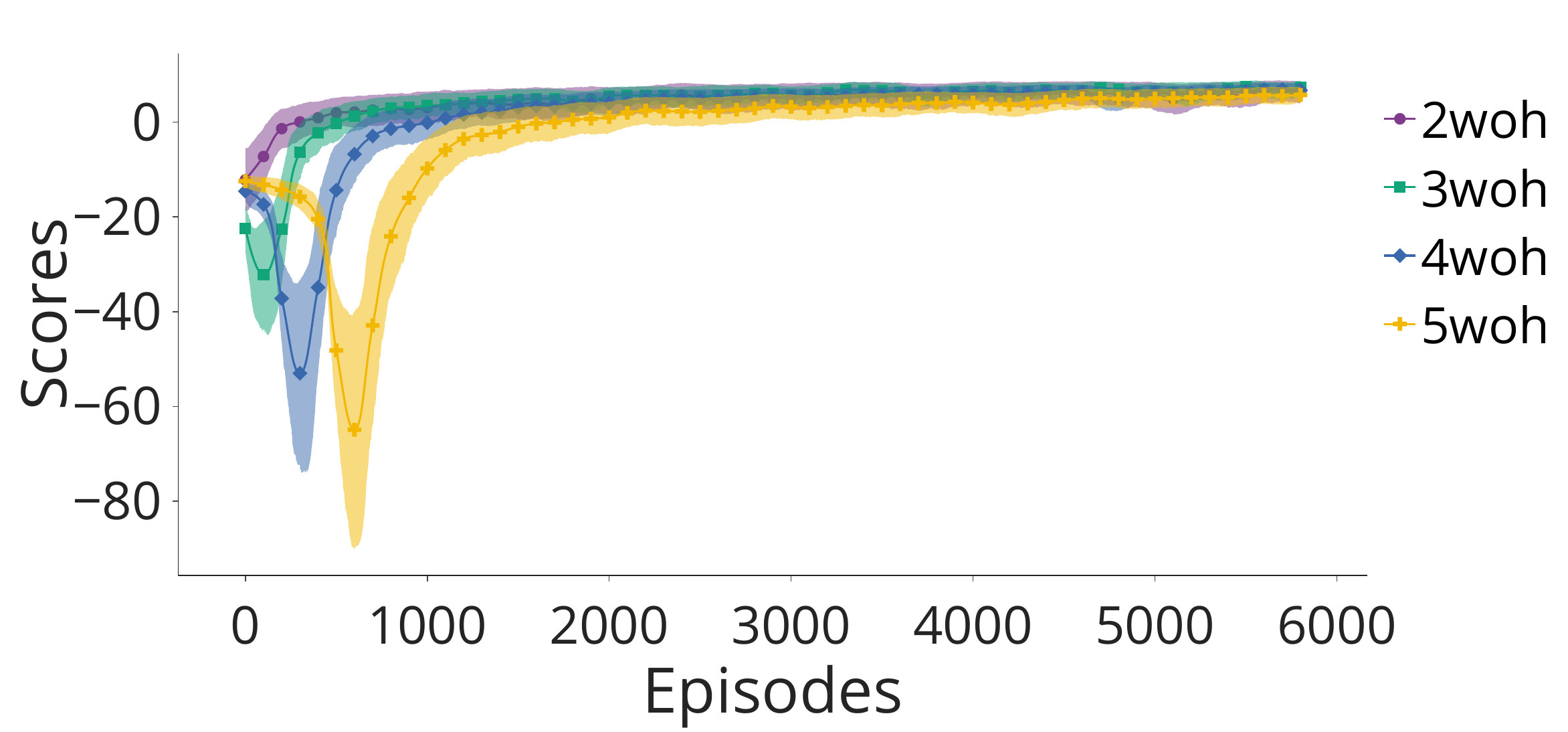}
  \caption{{The figure shows the performance of RPPO algorithm on scenario \texttt{Repairing a Flat Bicycle Tire} (without handicap) on multiple choice settings, 2, 3, 4, 5 respectively.}} 
  \label{fig:RPPO-2345actions-bicycle-no-hint}
\end{figure} 

\begin{figure}[t]
\centering
  \includegraphics[scale=0.32]{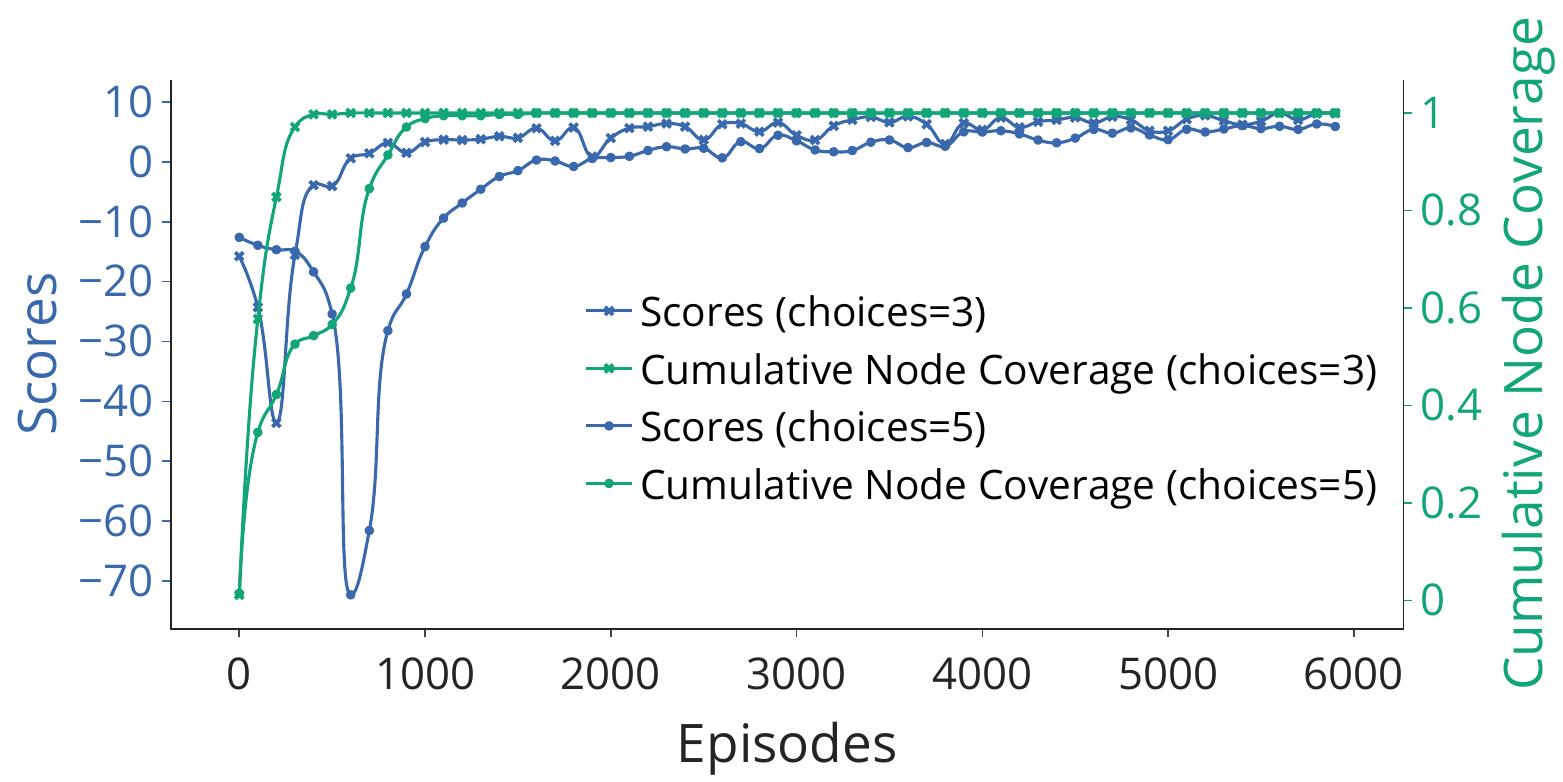}
  \caption{The figure shows the performance of RPPO algorithm on scenario \texttt{Repairing a Flat Bicycle Tire} (without handicap) on choices, 3 and, 5 with respective node coverages across learning. The increasing coverage slope (green) and the performance dip (blue) coincide in both settings highlighting the role of graph coverage in algorithm's learning.} 
  \label{fig:performance-dip-analysis}
\end{figure} 

\noindent\textbf{Generalization across Scenarios:} In \scriptworld\, since all the scenarios belong to real-life daily activities, an interesting experiment is to test the generalization capability of an algorithm trained on a specific scenario. We chose two similar (in terms of commonsense knowledge required to solve) scenarios, \texttt{Going on a Train} and \texttt{Riding on a Bus}, for this experiment. Table \ref{tab:generalization-train-test} shows the evaluation performance of RPPO on all scenarios trained on one scenario. We observe that the RPPO algorithm generalizes more across similar scenarios e.g., between \texttt{Train} and \texttt{Bus} (more details in App. D). 
Results obtained in this experiment also open up new research directions like test-time domain adaptation and continual learning.

\begin{table}[t]
\centering
\tiny
\renewcommand{\arraystretch}{1}
\setlength\tabcolsep{1pt}
\begin{tabular}{ccccccccccc}
\toprule
\multirow{2}{*}{\begin{tabular}[c]{@{}c@{}}Training\\ Scenario\end{tabular}} & \multicolumn{10}{c}{Performance on other Scenarios}                                  \\
\cmidrule{2-11}
                                & \texttt{Airplane} & \texttt{Bath} & \texttt{Bicycle} & \texttt{\textbf{Bus}}  & \texttt{Cake} & grocery & \texttt{Haircut} & \texttt{Library} & \texttt{\textbf{Train}} & \texttt{Tree} \\
                                \midrule
\texttt{\textbf{Bus}}                           & -24.78     & -15.07 & -5.02    & \textbf{9.97} & -23.39  & -26.51    & -20.85    & -16.29    & \color{red}\textbf{2.14}\color{black}  & -21.78  \\
\texttt{\textbf{Train}}                             & -11.31     & -13.22 & -9.52    & \color{red}\textbf{5.44}\color{black} & -4.42 & 0.22    & -10.97    & -6.79    & \textbf{9.56}   & -0.59\\
\bottomrule
\end{tabular}
\caption{The table shows performance on RPPO algorithm trained one scenario and evaluated on all scenarios. RPPO trained on \texttt{Bus} performs better on \texttt{Train} and vice versa (highlighted in \color{red}{red}\color{black}), depicting the generalization across scenarios.}
\label{tab:generalization-train-test}
\end{table}

\noindent\textbf{Performance on different choice settings:} To benchmark the flexibility feature of choosing the number of actions in the environment setting, we also report the results for RPPO on various numbers of actions. Fig. \ref{fig:RPPO-2345actions-bicycle-no-hint} shows the training curves for settings with choices = 2, 3, 4, 5, highlighting the increasing difficulty level as the number of choices in the environment increases. We observe an interesting trend, the occurrence of a performance dip in all the scenarios for different episode numbers. Notice the performance dip in Fig. \ref{fig:RPPO-2345actions-bicycle-no-hint} for all the runs with varying numbers of choices. As can be observed, the episode for performance dip increases with the increasing number of choices in the environment. We study this behavior of RL algorithms in detail by analyzing the trajectory followed by the RL algorithms. Fig. \ref{fig:performance-dip-analysis} shows the percentage coverage of scenario graph nodes along with rewards. The point for a maximum dip (after which the algorithm starts improving the score) directly coincides with the increasing percentage of node coverage; we speculate that the algorithm begins developing a mapping for each node after the entire graph exploration and works on improving the node representation in the later episodes. Though the graph coverage percentage is higher, it still remains a difficult task to optimize for correct choice as the number of paths in the graph is huge, and the choices generated for each node are random, making each scenario node different at different time steps. 

%% file: Discussion.tex
\section{Discussion and Future Directions} \label{sec:discussion}

\scriptworld\ provides a suitable benchmark to test different settings as it provides flexibility to adjust the game's complexity. The environment has certain limitations. For example, currently, the environment provides actions available at any state in the form of choices and does not allow the agent to generate actions in free-form text. 
This limitation is also there in the current parser based text-games that restrict the vocabulary size and sentence constructions that an agent can use for interaction. Parsing and understanding free-form text is a non-trivial task for the current state-of-the-art NLP technologies. 
In the future, we plan to develop a parser-based version (allowing free-form text) of the game, making use of LLMs. \scriptworld\ 's current version only has 10 scenarios. This is mainly due to limitations from the DeScript corpus. In future work, we will try to address this by including more daily scenarios. Experiments show that agents struggle in no handicap setting since they do not have any prior knowledge about the real world. It would be interesting to incorporate external knowledge into agents in the future and explore the possibility of including human feedback for learning a new scenario. Alternatively, another idea to explore would be to allow agents to gather information about a task from the internet via search or by probing large language models. Including multiple diverse scenarios in the proposed environment can facilitate the validation of generalization and language understanding capabilities in fields like continual learning, where a single algorithm learns various tasks without catastrophic forgetting \cite{nguyen2019toward}.


%% file: Conclusion.tex
\section{Conclusion} \label{sec:conclusion}

In this paper, we present a novel approach to building a text-based game environment (\scriptworld) involving different daily scenarios. This is a step towards training RL agents to develop NLU capabilities and commonsense knowledge about the real world. We perform an extensive set of experiments. Our experiments and analysis not only explore the environment in RL setting but also open up new ways in which the environment is helpful for the research community. 


%% file: Appendix.tex
\section*{Appendix}
\label{appendix}

\appendix

\renewcommand{\partname}{}{} 
\renewcommand{\thepart}{} 

\vspace{-15mm}

\addcontentsline{toc}{section}{Appendix} 
\part{} 
\parttoc 

\listofappendixfigures
\listofappendixtables

\newpage
\section{\scriptworld\ Details}\label{app:scriptworld-stats}


\subsection{\scriptworld\ Statistics and Comparison with Other Environments}

The statistics of the \scriptworld\ environment are provided in Table \ref{tab:stats}. Table \ref{tab:scenario_nodes_distance} shows the number of compact nodes in each scenario along with the negative sampling distance used for each scenario. 
The primary novelty of the created RL environment comes from the use of real-world collected scripts dataset in contrast to artificially made text-based games. We provide a brief feature comparison of the proposed \scriptworld\ environment with existing text-based environments in Table \ref{tab:comparison_with_other_environments}. 

\begin{table}[h]
\centering
\small
\renewcommand{\arraystretch}{1}
\setlength\tabcolsep{1pt}
\begin{tabular}{cccc}
\toprule
Stats                    & \scriptworld \\
\midrule
Vocabulary Size            &     1631         \\
Avg. Words / observation &     4.91        \\
\# States                &     456.4        \\
State Transitions       &    Deterministic          \\
Text Game Type           &    Choice based       \\
\bottomrule
\end{tabular}
\caption{Statistics of \scriptworld\ environment}
\label{tab:stats}
\end{table}

\begin{table}[h]
\small
\centering
\renewcommand{\arraystretch}{1}
\setlength\tabcolsep{1pt}
\begin{tabular}{cccc}
\toprule
Scenario\hspace{2mm}  & No. of compact Nodes &\hspace{2mm} Negative sampling distance  \begin{tabular}[c]{@{}c@{}}  \end{tabular} &              \\
\midrule
Bath                     & 32   & 7                                                          \\
Cake                     & 29   & 6                                                         \\
Airplane            & 36   & 8                                                        \\
Shopping            & 32   & 8                                                         \\
Train                  & 25   & 6                                                         \\
Tree                   & 24   & 5                                                          \\
Bus                   & 21   & 5                                                         \\
Bicycle     & 28   & 6                                                          \\
Library & 22   & 5                                                          \\
Haircut                & 39   & 9                                                         \\

\bottomrule
\end{tabular}
\caption{The table contains the number of nodes in compact graphs for different scenarios and negative sampling distance (minimum  distance either forward or backward from where wrong actions will be sampled).}
\label{tab:scenario_nodes_distance}
\end{table}

\begin{table}[h]
\centering
\tiny
\renewcommand{\arraystretch}{1}
\setlength\tabcolsep{5pt}
\begin{tabular}{ccc}
\toprule
\textbf{Environments}  & \textbf{Game Type (basis)}           & \textbf{Modalities} \\
\midrule
Text World             & Parser/ Choice  & Text                \\
TWC & Parser                  & Text                \\
Jericho                & Template                & Text                \\
Pyfiction              & Hypertext               & Text                \\
Nethack*               & Visual and Text         & Visual, Text        \\
3D World*              & Choice                  & Visual, Text        \\
VizDoom*               & Choice                  & Visual, Text        \\
QAit                   & Parser                  & Text                \\
Evennia                & Parser                  & Text                \\
\textbf{\scriptworld\ (ours)}    & \textbf{Choice}                  & \textbf{Text}               \\
\bottomrule
\end{tabular}
\caption{The table shows a comparison between different existing text-based environments and \scriptworld.}
\label{tab:comparison_with_other_environments}
\end{table}


\subsection{Compact to Scenario Graph Creation} 

\begin{figure}[t]
\centering
  \includegraphics[width=\linewidth]{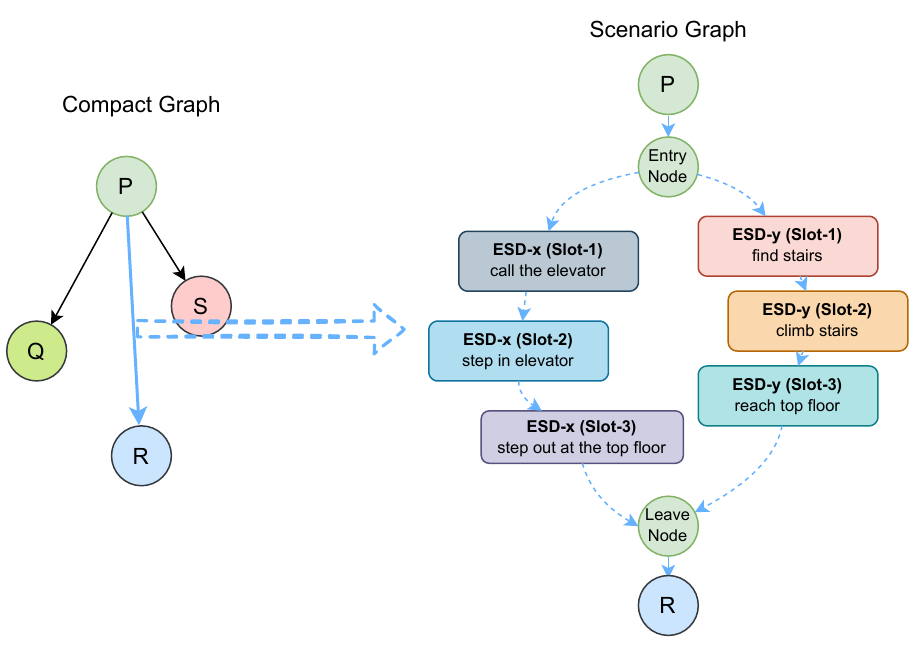}
  \caption[Compact to Scenario Graph creation]{{Figure shows a dummy example of a compact graph split into scenario graph.}}
  \label{fig:compact_graph_entry_exit}
\end{figure}

Fig. \ref{fig:compact_graph_entry_exit} shows an example of a compact graph with entry and exit nodes (\S\ref{sec:scr-env}). The motivation for creating the compact graph was due to the type of annotations provided in the “DeScript”-dataset. “DeScript” provides annotations in two aspects 1) the annotations group the actions belonging to a specific event together (compact graph’s node in our case) and 2) creating the alignment of these events for multiple scripts. Hence, we used the provided alignments to create a compact graph first and later to consider the multiple sub-paths to enrich it further(calling it a scenario graph). Moreover, another use of compact graphs would be in future work, as the RL algorithms can be trained to learn specific sub-tasks. E.g., if multiple scenarios have the same sub-tasks(events or sequence of events), the same knowledge can be transferred across multiple scenarios like \texttt{Riding a Bus} or \texttt{Riding a Train}. Also, the environment’s reward can be set in such a way that rewards more on learning the generic events rather than the scenario-specific events.

\subsection{Graphs for Scenarios}\label{app:compact-graphs}

\begin{figure*}[t]
\centering
  \includegraphics[scale=0.80]{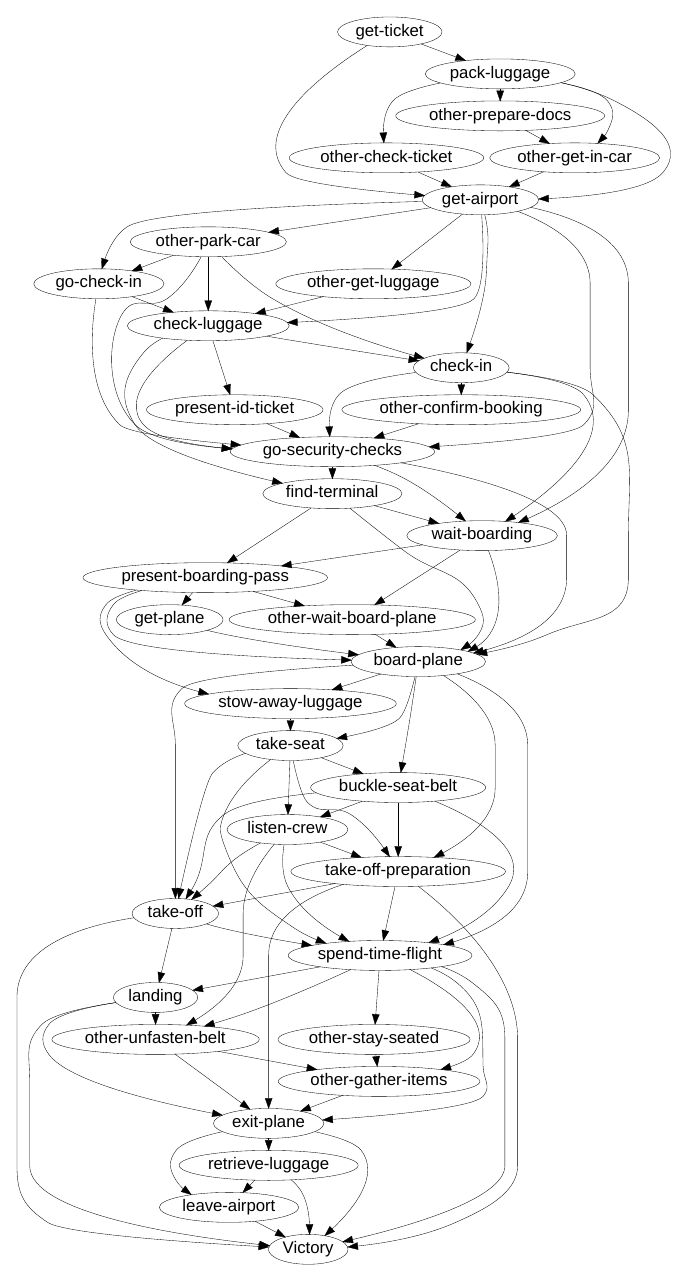}
  \caption[Compact Graph ``airplane"]{{The figure shows the compact graph created for the scenario \texttt{Flying in an Airplane} }}
  \label{fig:airplane_compact_graph}
\end{figure*}

\begin{figure*}[t]
\centering
  \includegraphics[scale=0.80]{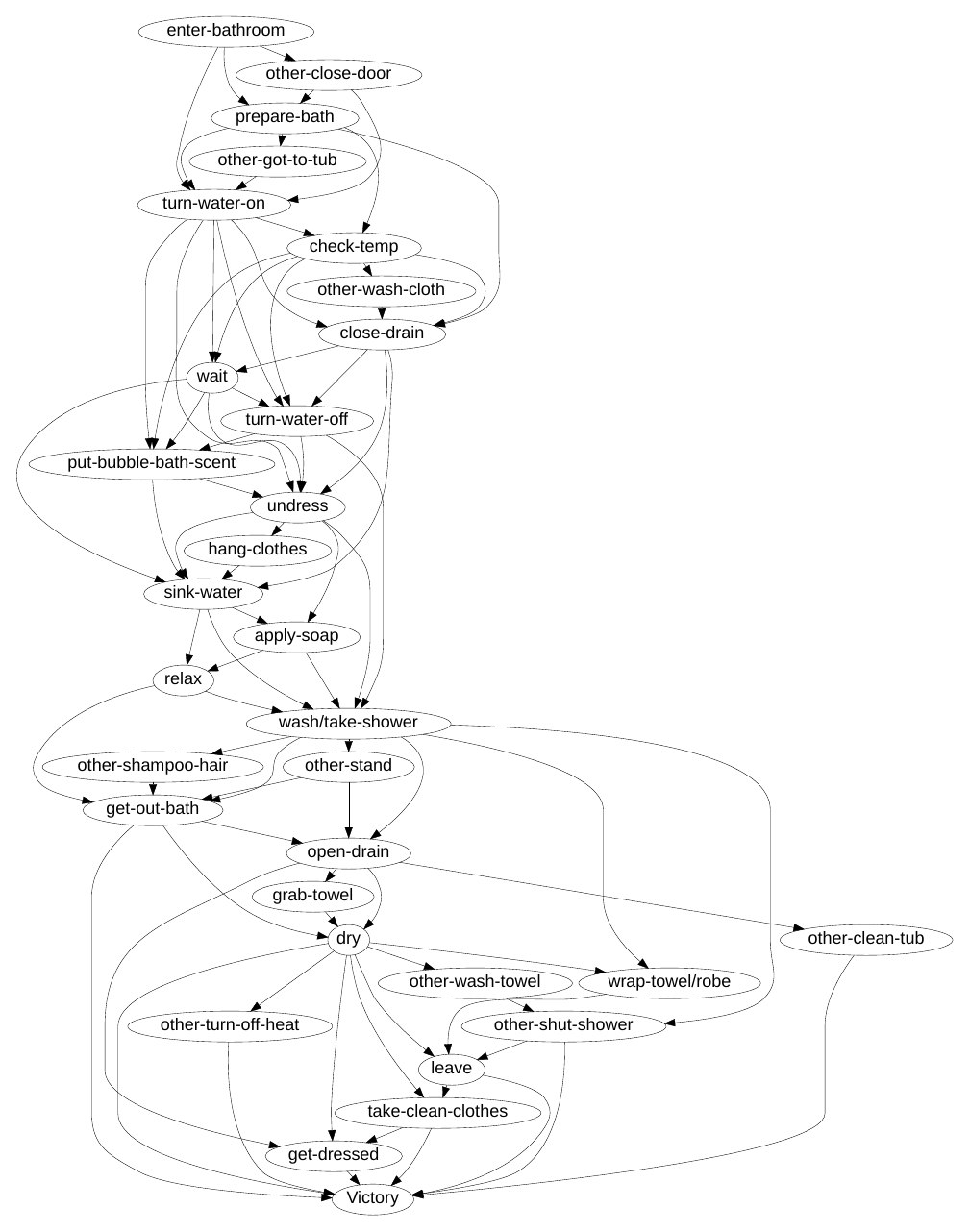}
  \caption[Compact Graph ``bath"]{{The figure shows the compact graph created for the scenario \texttt{Taking a Bath} }}
  \label{fig:bath_compact_graph}
\end{figure*}

\begin{figure*}[t]
\centering
  \includegraphics[scale=0.70]{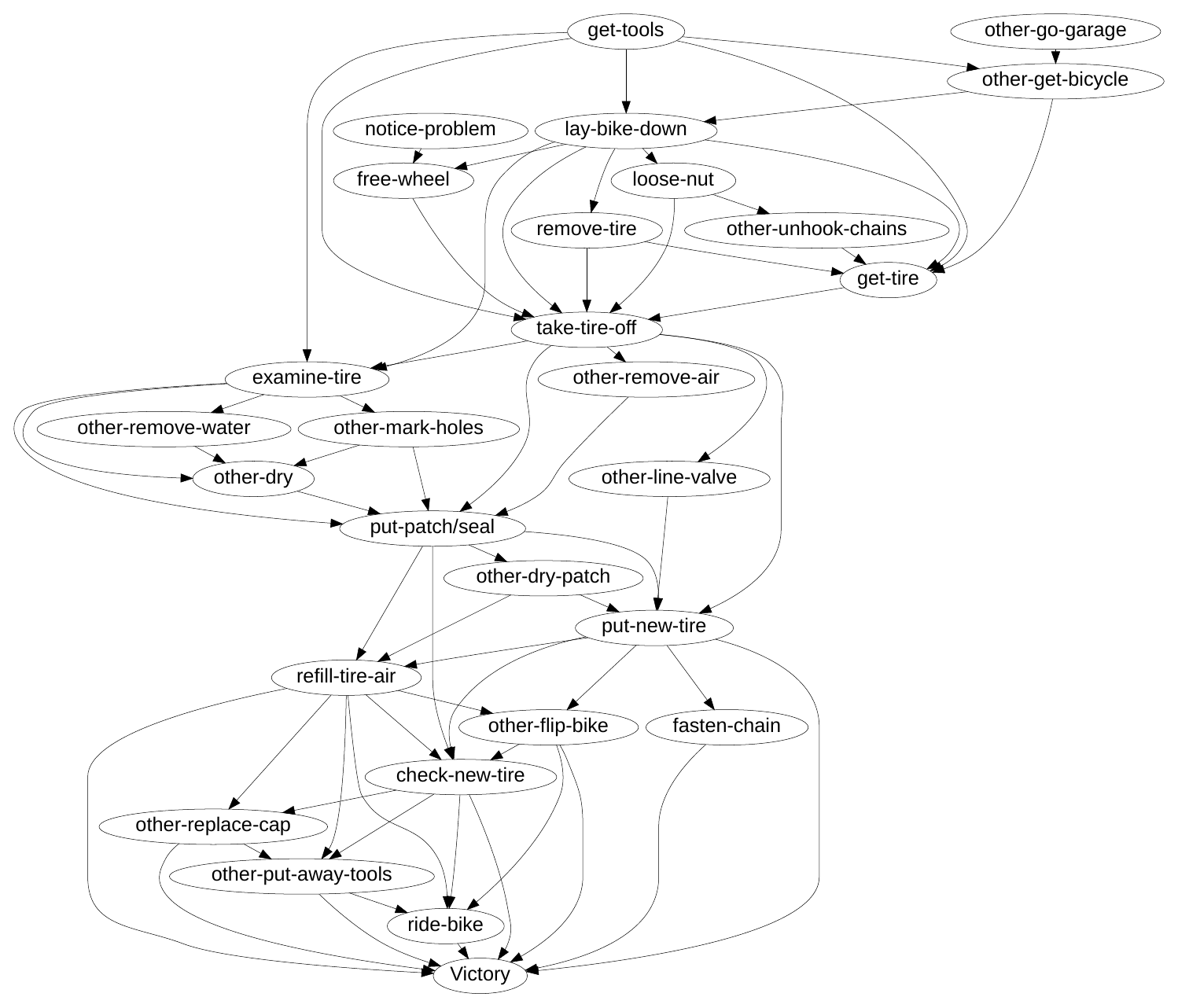}
  \caption[Compact Graph ``bicycle"]{{The figure shows the compact graph created for the scenario \texttt{Repairing a Flat Bicycle Tire} }}
  \label{fig:bicycle_compact_graph}
\end{figure*}

\begin{figure*}[t]
\centering
  \includegraphics[scale=0.80]{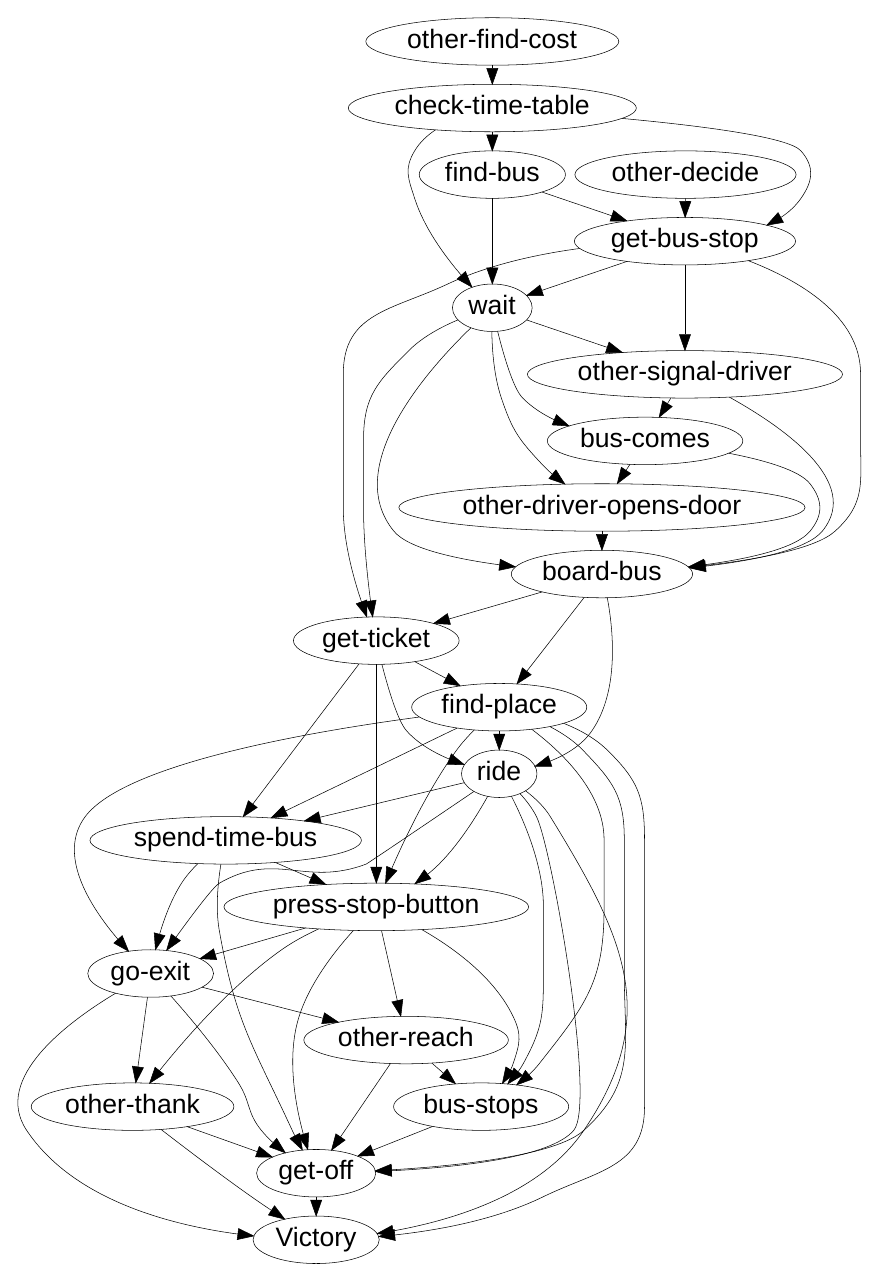}
  \caption[Compact Graph ``bus"]{{The figure shows the compact graph created for the scenario \texttt{Riding on a Bus} }}
  \label{fig:bus_compact_graph}
\end{figure*}

\begin{figure*}[t]
\centering
  \includegraphics[scale=0.80]{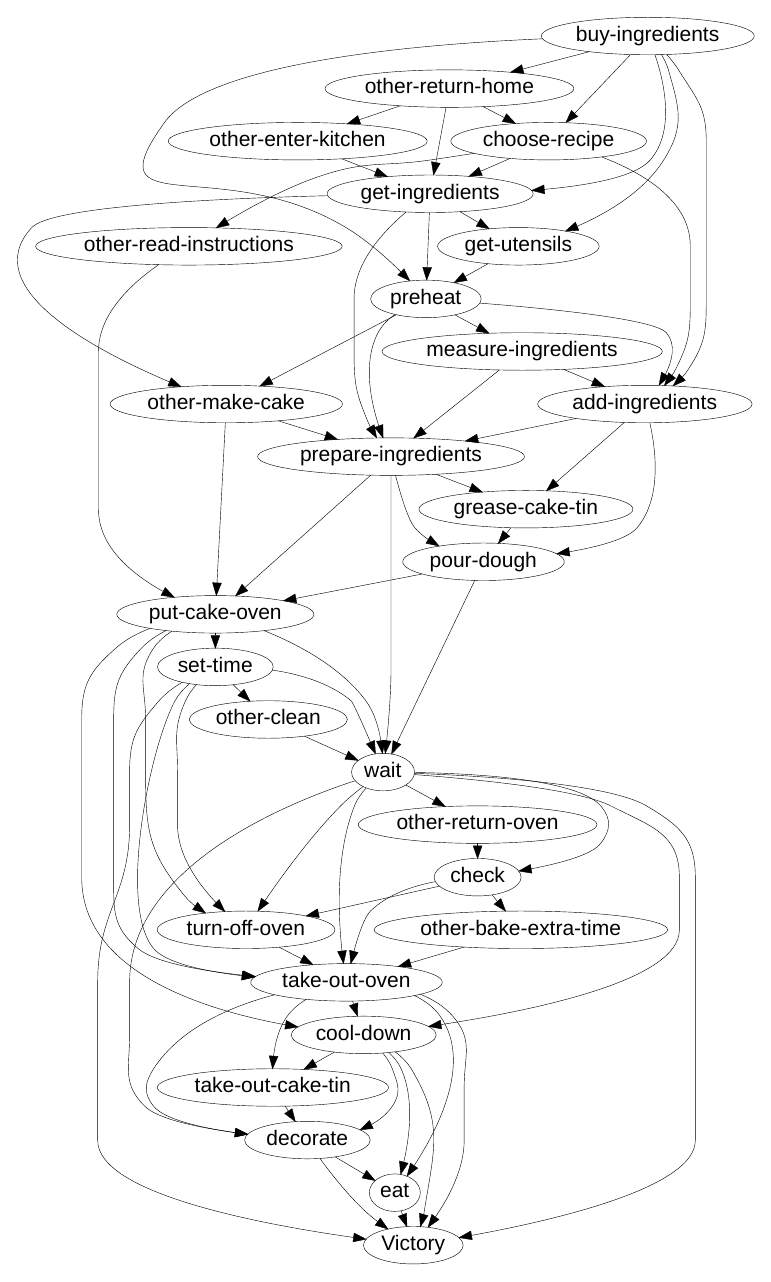}
  \caption[Compact Graph ``cake"]{{The figure shows the compact graph created for the scenario \texttt{Baking a Cake} }}
  \label{fig:cake_compact_graph}
\end{figure*}

\begin{figure*}[t]
\centering
  \includegraphics[scale=0.80]{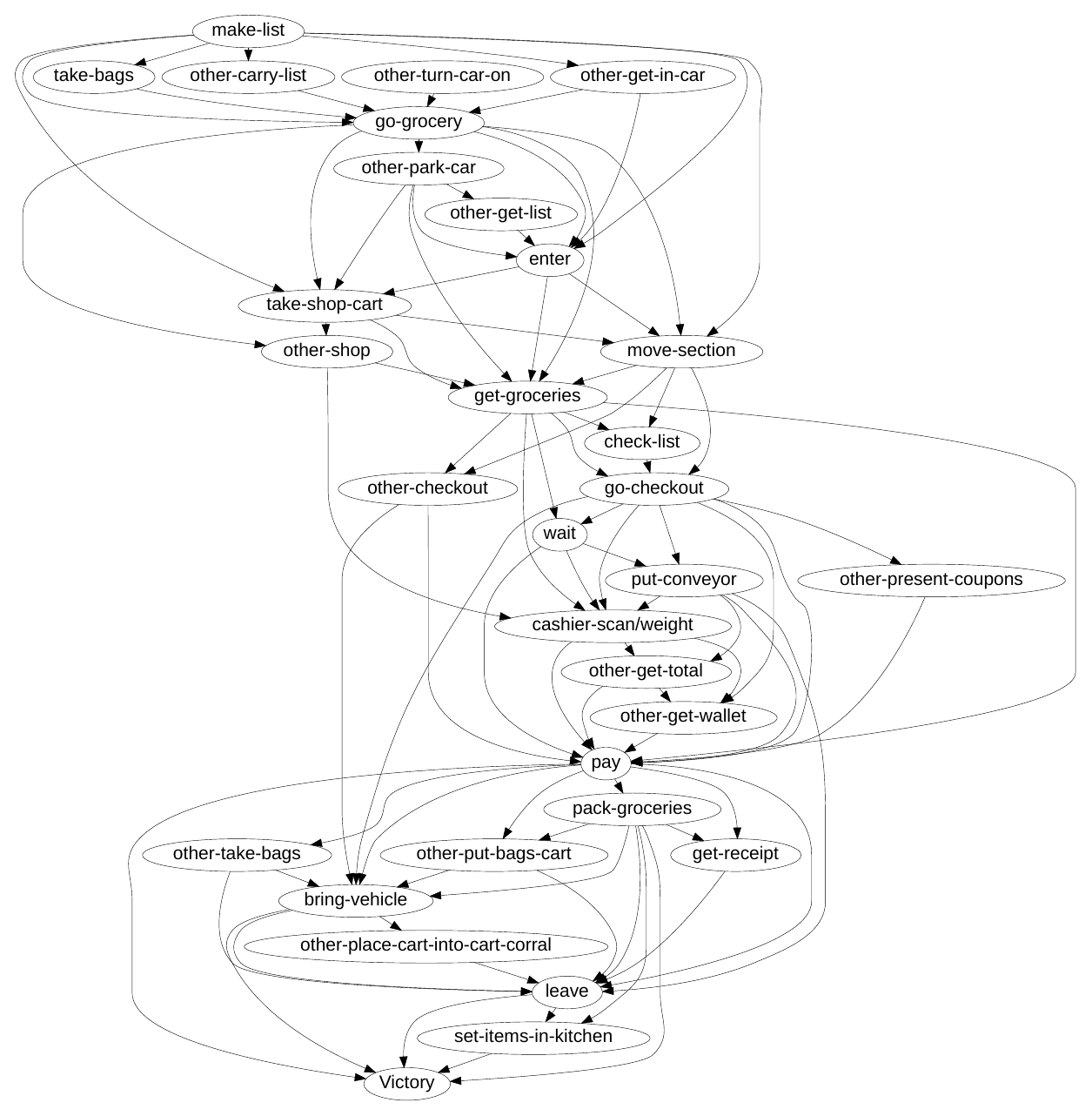}
  \caption[Compact Graph ``shopping"]{{The figure shows the compact graph created for the scenario \texttt{Going Grocery Shopping} }}
  \label{fig:shopping_compact_graph}
\end{figure*}

\begin{figure*}[t]
\centering
  \includegraphics[scale=0.80]{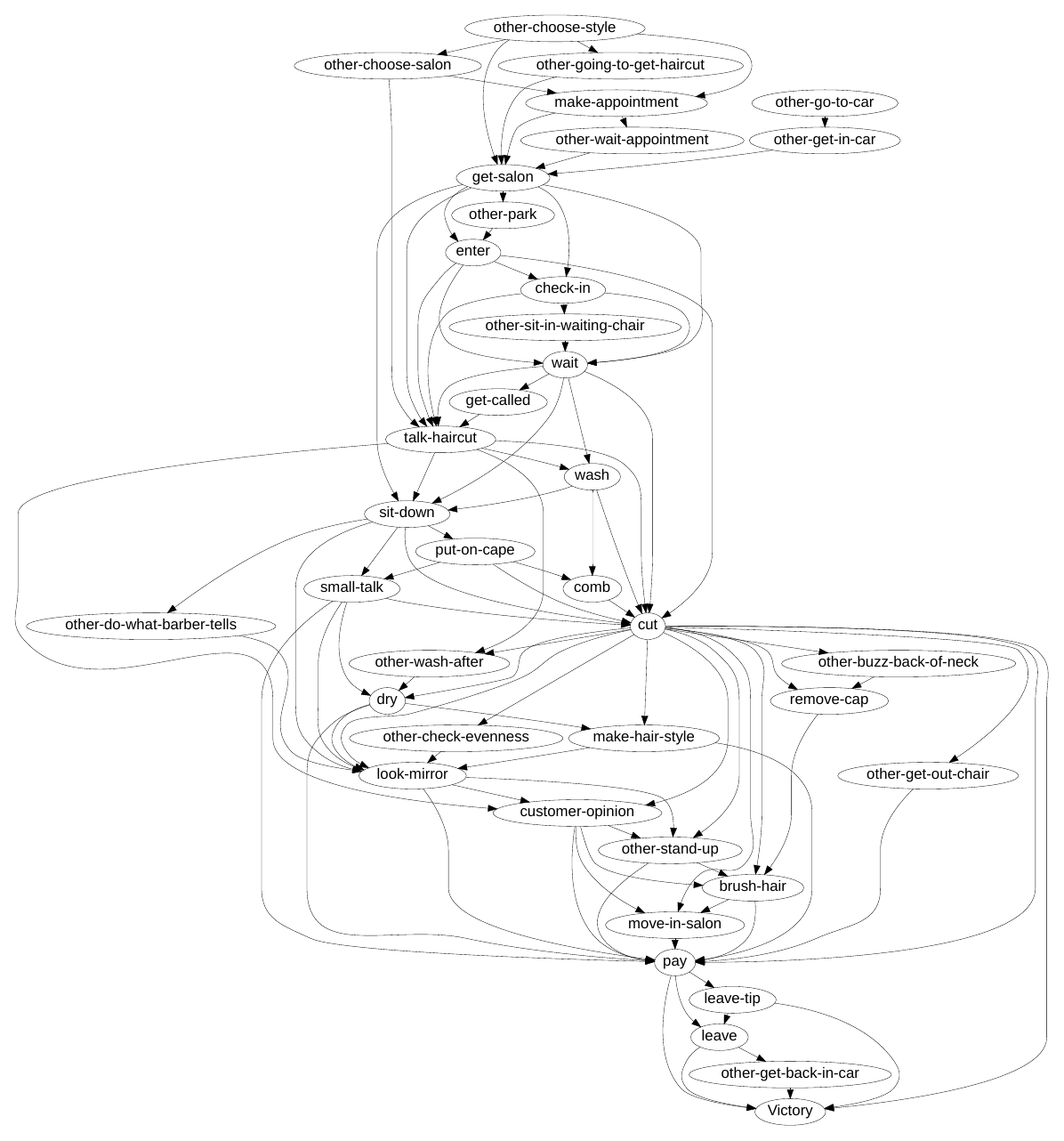}
  \caption[Compact Graph ``haircut"]{{The figure shows the compact graph created for the scenario \texttt{Getting a Haircut} }}
  \label{fig:hair_compact_graph}
\end{figure*}

\begin{figure*}[t]
\centering
  \includegraphics[scale=0.80]{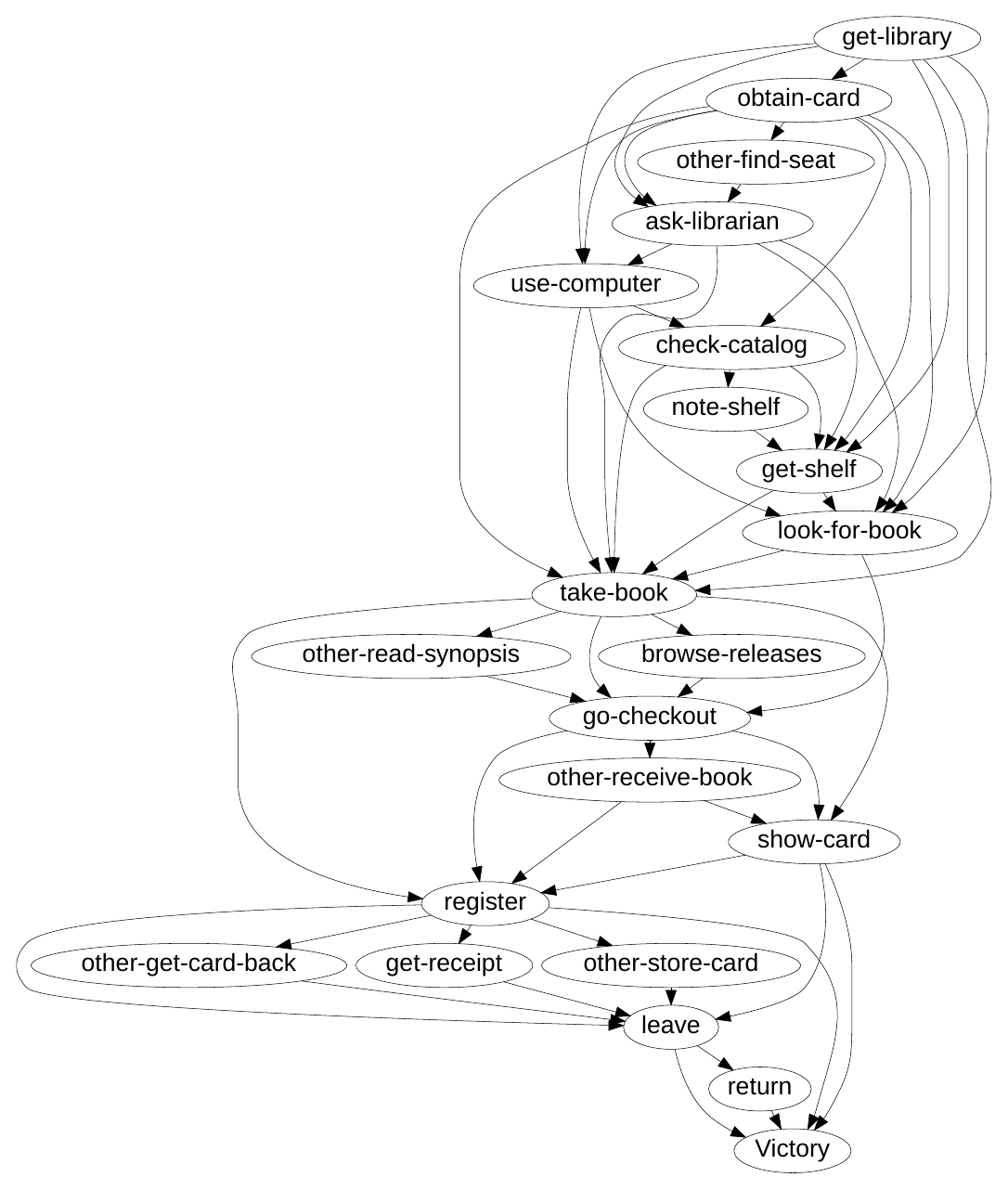}
  \caption[Compact Graph ``library"]{{The figure shows the compact graph created for the scenario \texttt{Borrowing a Book from the Library} }}
  \label{fig:library_compact_graph}
\end{figure*}

\begin{figure*}[t]
\centering
  \includegraphics[scale=0.80]{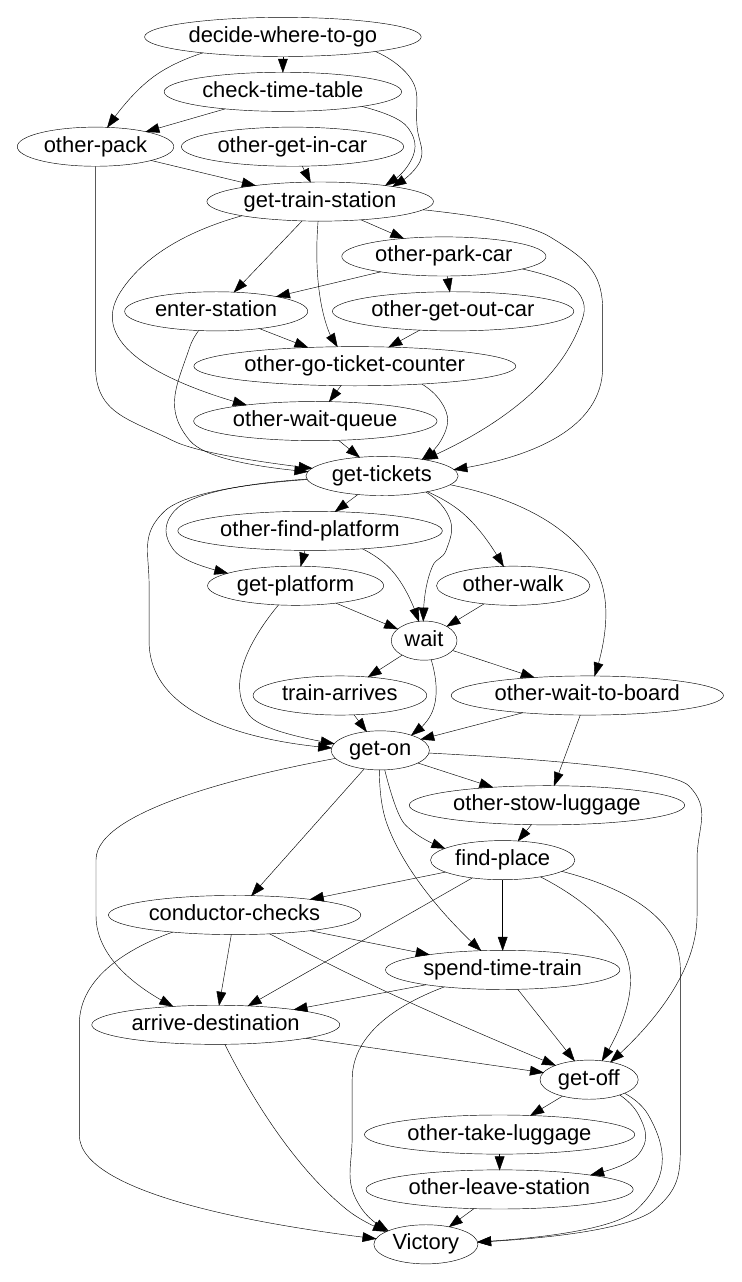}
  \caption[Compact Graph ``train"]{{The figure shows the compact graph created for the scenario \texttt{Going on a Train} }}
  \label{fig:train_compact_graph}
\end{figure*}

\begin{figure*}[t]
\centering
  \includegraphics[scale=0.80]{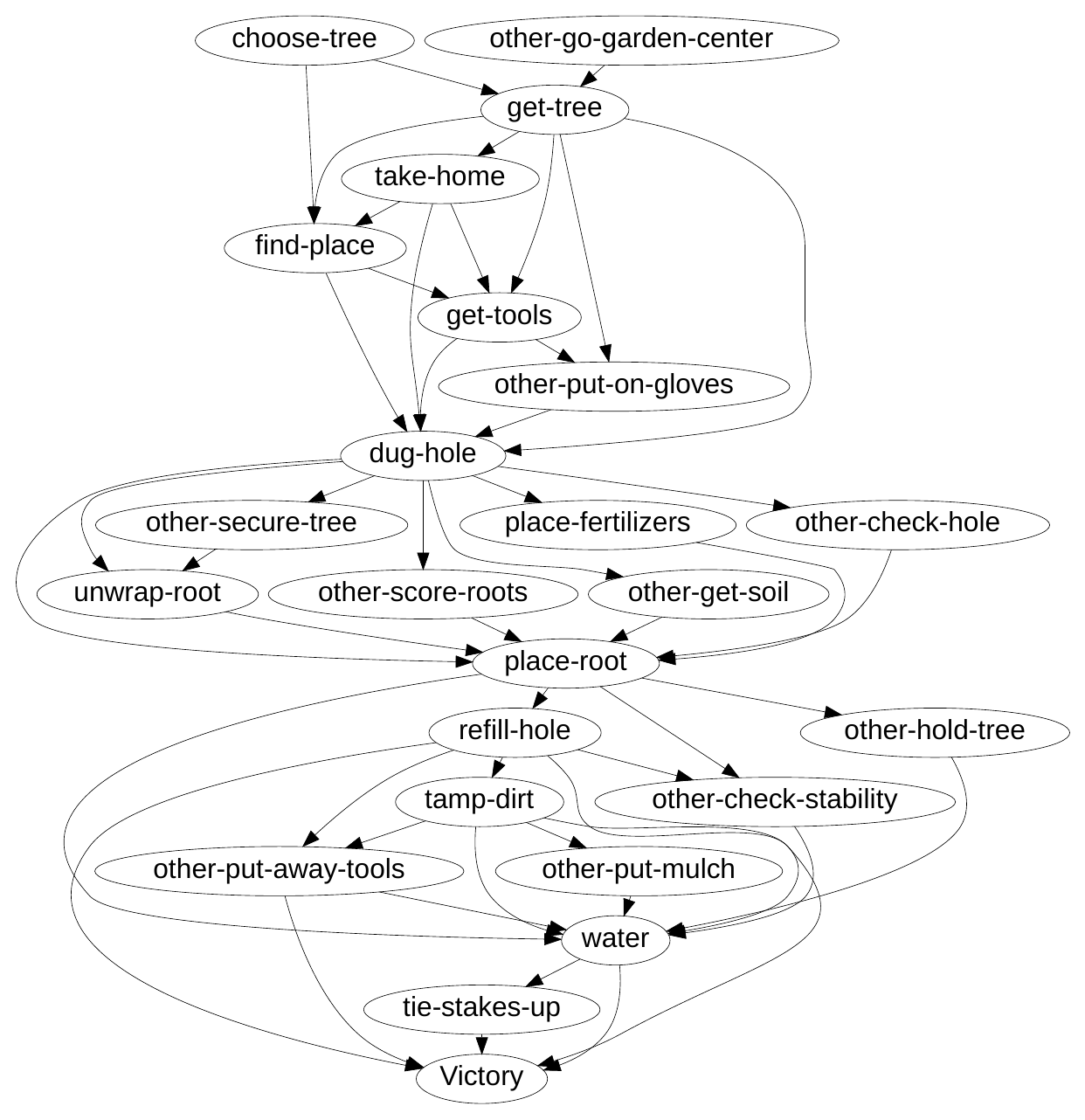}
  \caption[Compact Graph ``tree"]{{The figure shows the compact graph created for the scenario \texttt{Planting a Tree} }}
  \label{fig:tree_compact_graph}
\end{figure*}

The compact graphs for different scenarios are shown in
Figures \ref{fig:bath_compact_graph}, \ref{fig:cake_compact_graph},
\ref{fig:airplane_compact_graph}, \ref{fig:shopping_compact_graph}, \ref{fig:train_compact_graph}, \ref{fig:tree_compact_graph},  \ref{fig:bus_compact_graph}, 
\ref{fig:bicycle_compact_graph}, \ref{fig:library_compact_graph}, and 
\ref{fig:hair_compact_graph}.



\subsection{\scriptworld\ Game-play examples}\label{app:game-play}

\begin{figure*}[t]
\centering
  \includegraphics[scale=0.43]{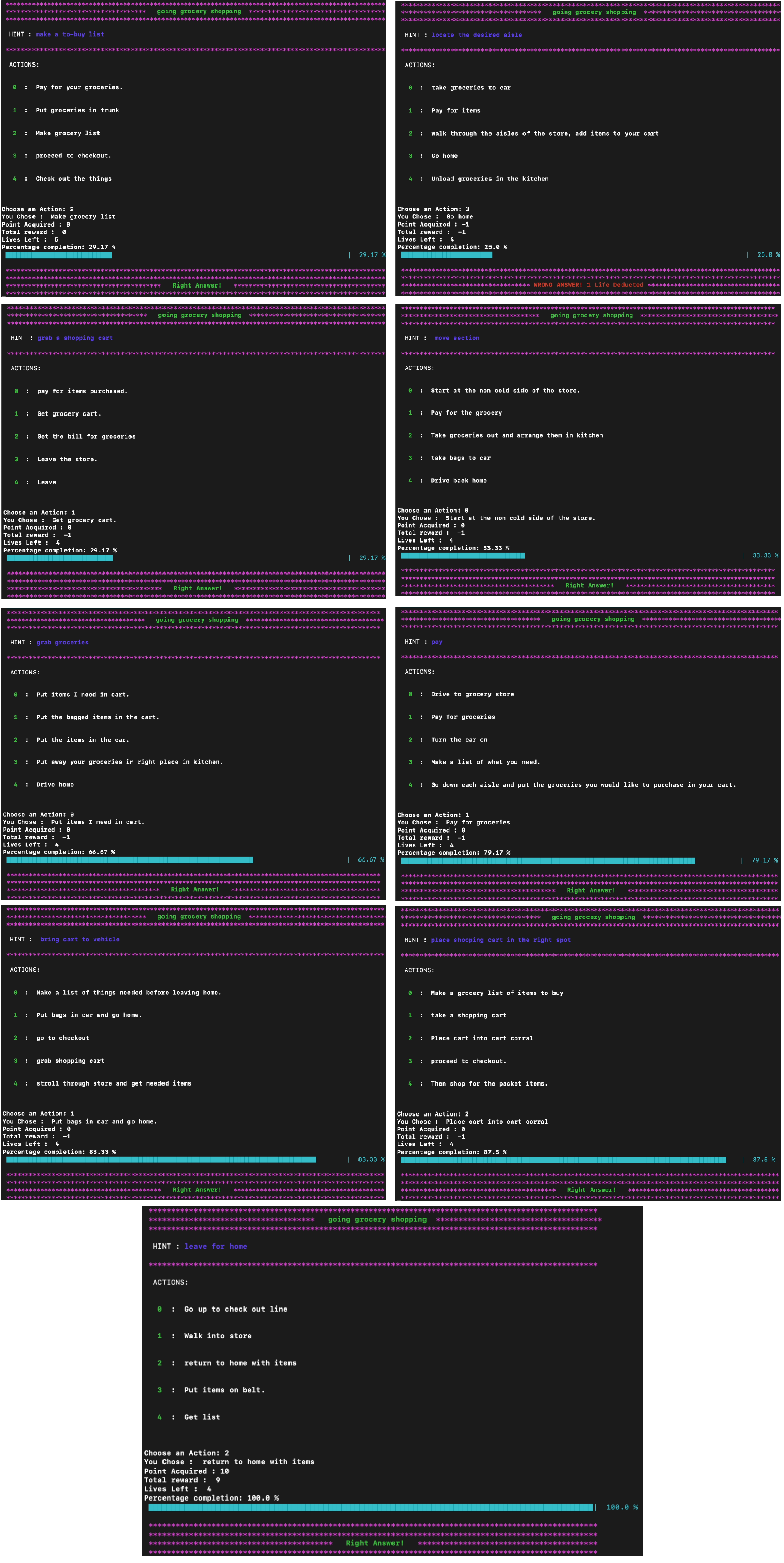}
  \caption[Gameplay ``shopping" (handicap, choices=5)]{{The figure shows a sample game-play (with hint)  for scenario \texttt{Going Grocery Shopping}. (the game-play sequences are left to right and top to bottom.) }}
  \label{fig:grocery-game-play-with-hint}
\end{figure*}

\begin{figure*}[t]
\centering
  \includegraphics[scale=0.33]{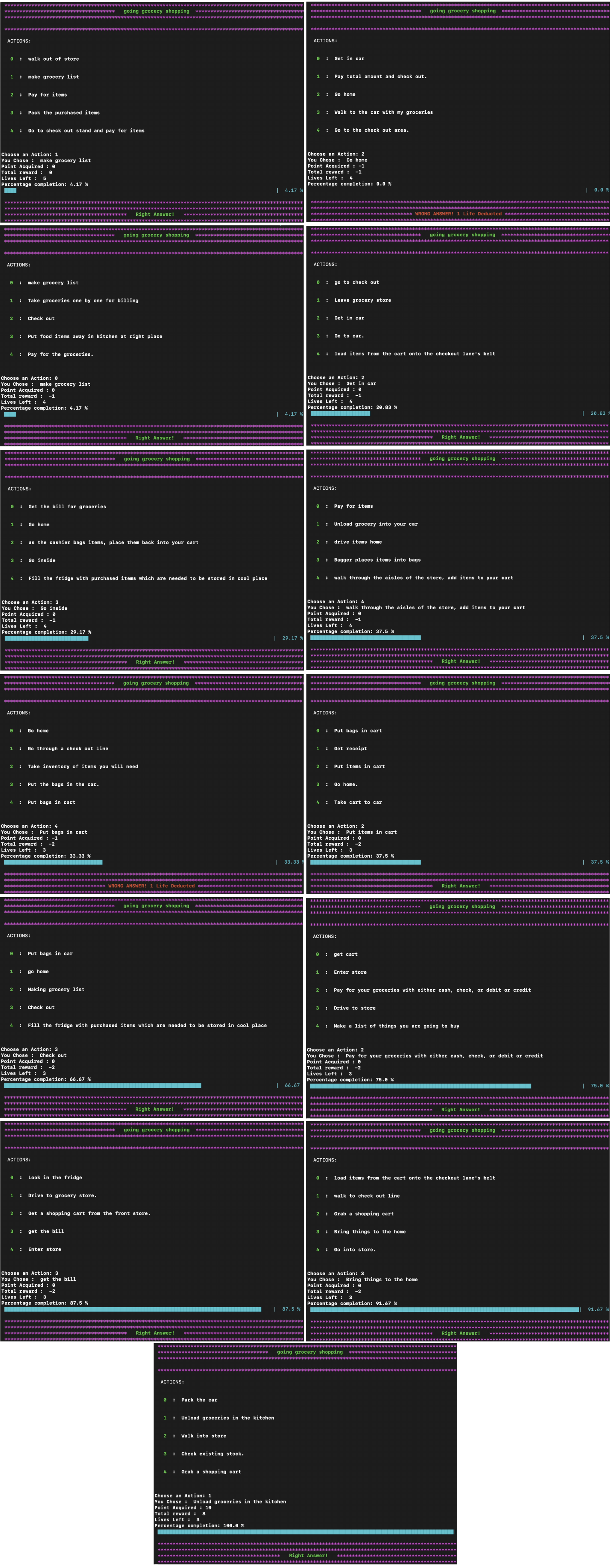}
  \caption[Gameplay ``shopping" (w/o-handicap, choices=5)]{{The figure shows a sample game-play (without hint) for scenario \texttt{Going Grocery Shopping}. (the game-play sequences are left to right and top to bottom.) }}
  \label{fig:grocery-game-play-without-hint}
\end{figure*}

Figures 
\ref{fig:grocery-game-play-with-hint} and \ref{fig:grocery-game-play-without-hint} show sample game play for the scenarion \texttt{Going Grocery Shopping}. 


\section{RL Algorithms Details}\label{app:rl-agents-details}


\subsection{RL agent Details}\label{app:q_learning_framework}
We explore multiple popular RL algorithms for baseline experiments. We formulate the RL framework by combining SBERT with RL algorithms. The observation-specific features are generated using a pre-trained SBERT model and used as input to the RL algorithms. We stick to the official implementations of StableBaselines3\footnote{\url{https://github.com/DLR-RM/stable-baselines3}} as our codebase and build upon it to benchmark the proposed \scriptworld\ environment.

\noindent\textbf{DQN:} DQN algorithm \cite{mnih2013playing} is trained to learn the action selection policies directly from the SBERT dimensional input. We use MLP (multilayer perceptron) variant in the DQN architecture and finetune the model parameters by manual inspections. 

\noindent\textbf{A2C:} A2C (Advantage Actor-Critic) \cite{mnih2016asynchronous_A2C_A3C} is a synchronous version of the A3C policy gradient method, which uses gradient descent for the optimization of feedforward parameters. A2C being synchronous, averages over all the actors for updates after the actors finish the segment.

\noindent\textbf{PPO:} PPO (Proximal Policy Optimization) \cite{PPO} uses trust regions to improve actors' performance, split the optimizations to multiple workers, like in A2C, and introduces clipping to avoid large updates keeping updated policies to be closer to the old policies.

\noindent\textbf{R-PPO:} RPPO (Recurrent-PPO) is the recurrent version of the PPO \cite{PPO} algorithm that combines recurrent policies with the help of LSTM layers keeping the rest of the algorithm the same. As the \scriptworld\ environment requires a continuous flow of several subtasks/actions for completing the required tasks, the recurrent nature of RPPO helps perform better in the environment.




\section{Evaluation Metrics}
\label{app:evaluation_metrics}

We use standard reward/scores vs. episodes as evaluation metrics for comparing the performance of multiple RL algorithms. All the results are averaged over 5 training iterations, with mean and standard deviations plotted on the plots.

For the generalization experiment \ref{sec:experiments} (Generalization across Scenarios), we train on one scenario and infer on all the scenarios by averaging performance over 100 runs. Fig. \ref{fig:RPPO_train_vs_all_error_plot} shows the average performance with standard deviations plotted as an error plot. Fig. \ref{fig:hm_2wh_RPPO}, Fig. \ref{fig:hm_2woh_RPPO}, Fig.  \ref{fig:hm_5wh_RPPO}, and Fig. \ref{fig:hm_5woh_RPPO} shows average performance scores plotted in a heatmap for a clear representation. 


\section{Additional Results And Analysis}\label{app:additional_results_analysis}


\subsection{Additional Experiments with Environment} \label{app:more-exp}
\begin{figure}[t]
\centering
  \includegraphics[width=\linewidth]{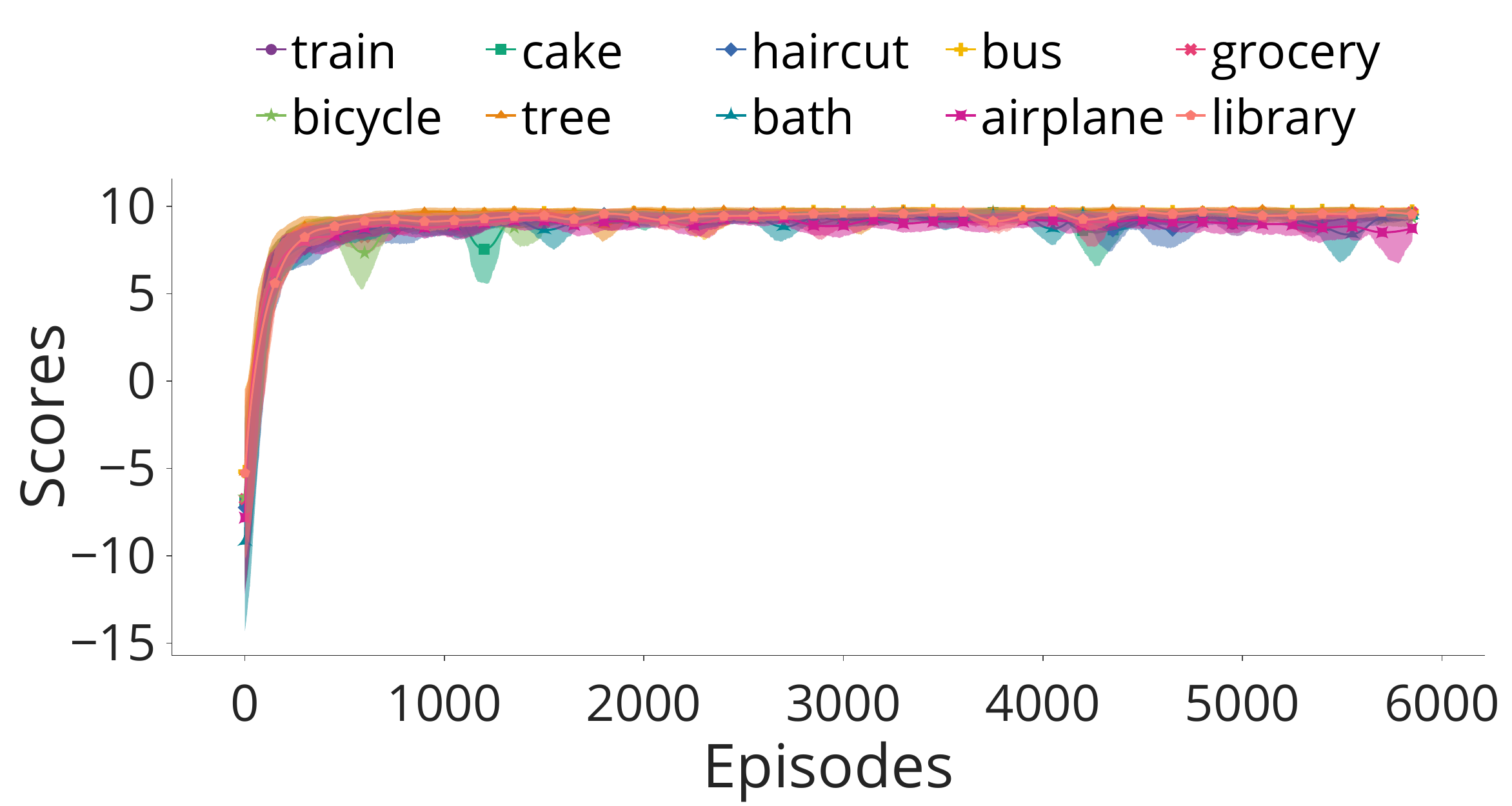}
  \caption[RPPO All scenarios (handicap, choices=2)]{{The figure shows the performance of RPPO algorithm on all scenarios with setting (with handicap, choices=2).}}
  \label{fig:All_scn_2wh_RPPO}
\end{figure}

\begin{figure}[t]
\centering
  \includegraphics[width=\linewidth]{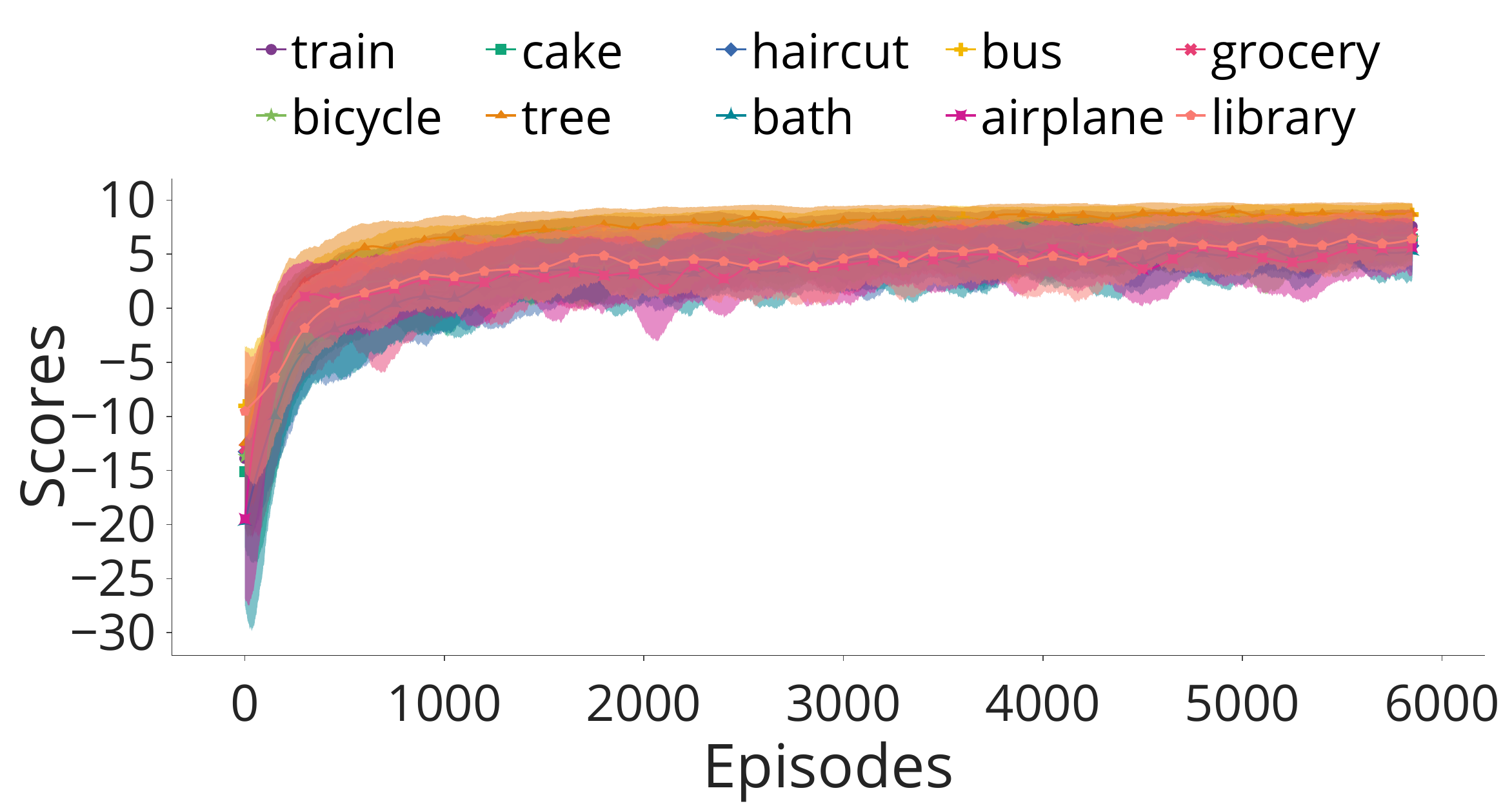}
  \caption[RPPO All scenarios (w/o handicap, choices=2)]{{The figure shows the performance of RPPO algorithm on all scenarios with setting (without handicap, choices=2). 
  }}
  \label{fig:All_scn_2woh_RPPO}
\end{figure}

To examine the performance of the best-performing RL algorithm RPPO on \scriptworld\ environment, we report the results of RPPO in various settings. Fig. \ref{fig:All_scn_2wh_RPPO} and Fig. \ref{fig:All_scn_2woh_RPPO} show a comparison of learning curves for all the scenarios in \scriptworld on with and without handicap settings, respectively. The setting without handicaps shows higher variance and poor performance than the one with the handicap. 

\begin{figure}[t]
\centering
  \includegraphics[width=\linewidth]{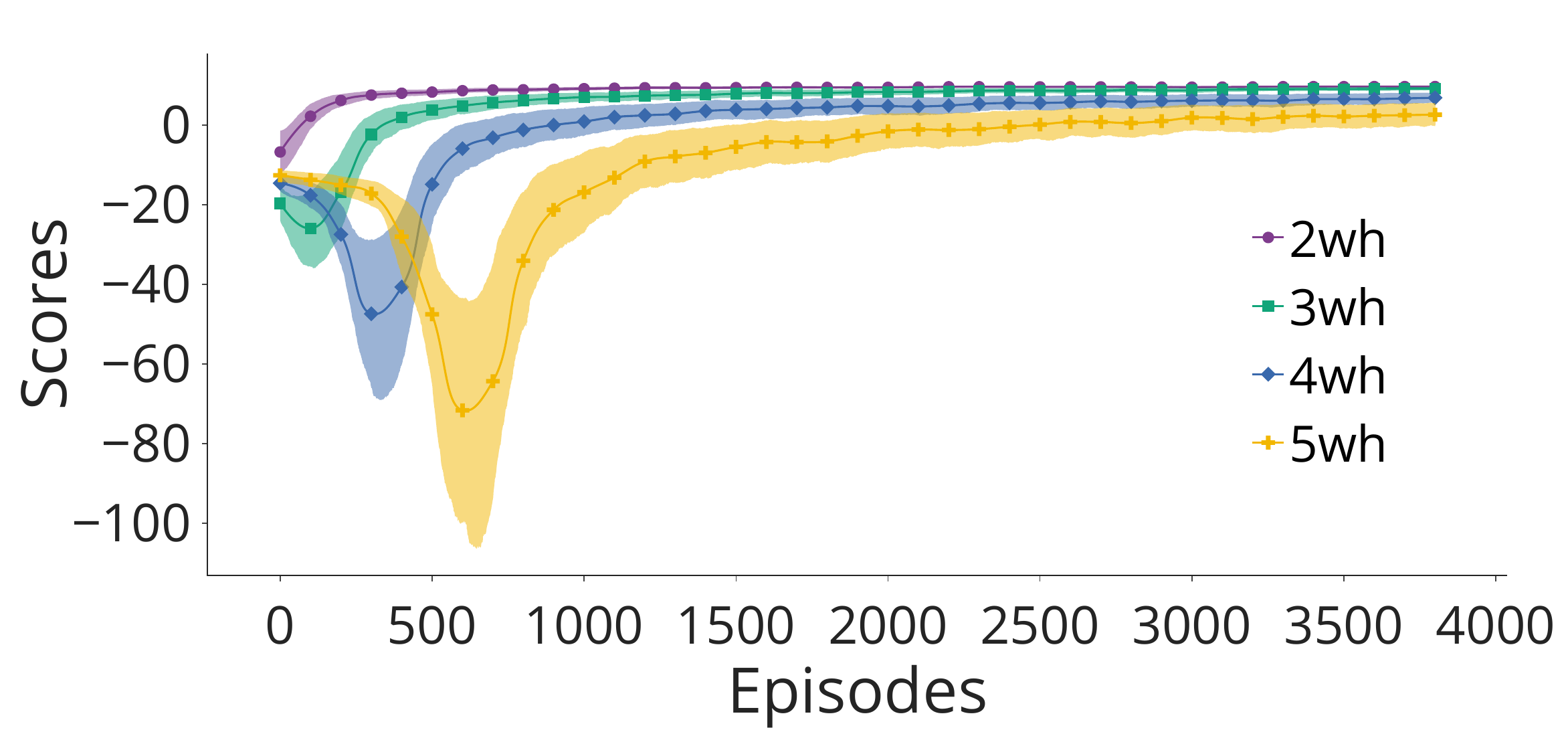}
  \caption[PPO (handicap, choices 2, 3, 4, 5)]{{The figure shows the performance of PPO algorithm on scenario \texttt{Repairing a Flat Bicycle Tire} (without handicap) on multiple-choice settings, 2, 3, 4, 5 respectively. The plot shows the increasing level of difficulty in the environment with an increase in the number of choices provided to the agent.}}
  \label{fig:PPO-2345actions-bicycle-withouthint}
\end{figure}

We also report additional results on the effect of increasing the number of choices in environment settings for other RL algorithms. Fig. \ref{fig:PPO-2345actions-bicycle-withouthint} shows the performance of the PPO algorithm in the scenario \texttt{Repairing a Flat Bicycle Tire}. The plot shows a similar trend as reported in the main paper for RPPO, with a high correlation between the increasing level of difficulty in the environment and the number of choices provided to the agent. We also observe a similar performance dip trend as observed in RPPO. (performance dip analysis and details in the Section \ref{sec:experiments})


\subsection{Performance with various LMs}\label{app:additional_LM_results}
\begin{figure}[t]
\centering
  \includegraphics[width=\linewidth]{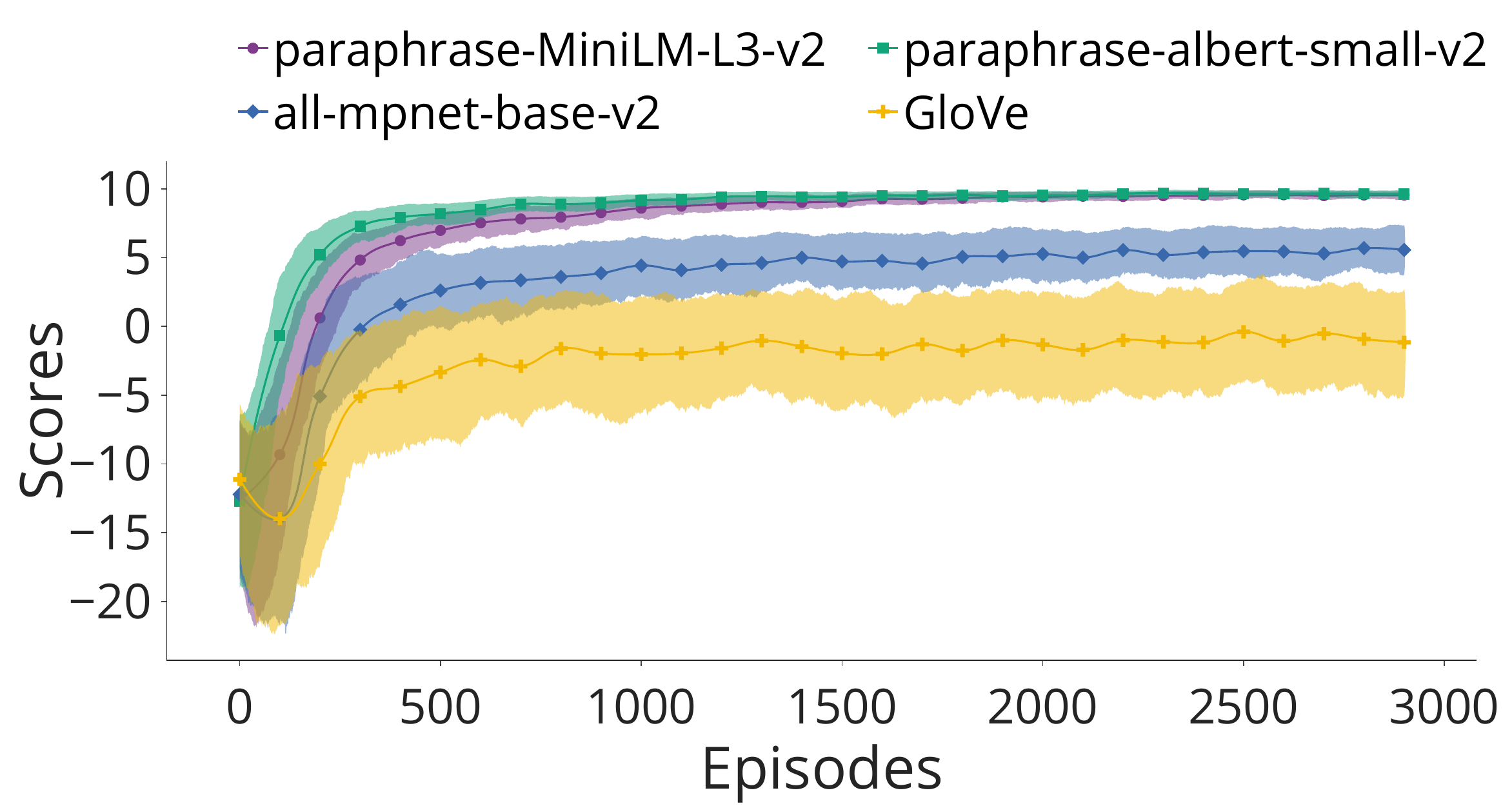}
  \caption[PPO diff. LM embds. (handicap, choices=2)]{{The figure shows the performance of PPO algorithm with various language models on scenario \texttt{Repairing a Flat Bicycle Tire} on setting (with a handicap, choices=2).}}
  \label{fig:LM_PPO_2wh_bicycle}
\end{figure}

\begin{figure}[t!]
\centering
  \includegraphics[width=\linewidth]{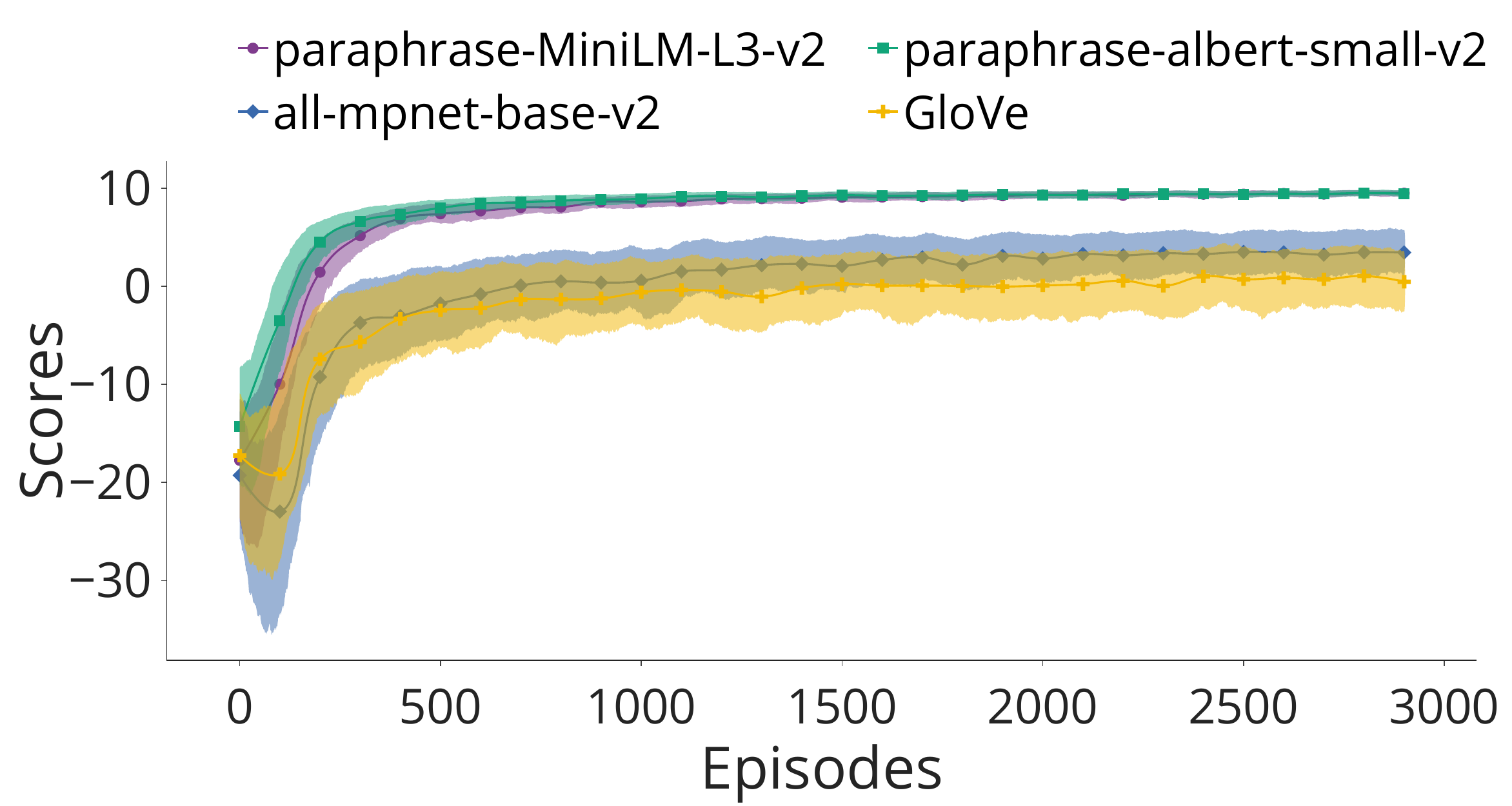}
  \caption[PPO diff. LM embeds. (handicap, choices=2)]{{The figure shows the performance of PPO algorithm with various language models on scenario \texttt{Taking a Bath} on setting (with handicap, choices=2).}}
  \label{fig:LM_PPO_2wh_bath}
\end{figure}

To explore the effect of various language models, we do an extensive set of experiments with multiple SBRT embeddings. Overall, language model embeddings perform better than the GloVe(non-contextual) embeddings. Fig. \ref{fig:LM_RPPO_2wh_bicycle} and Fig. \ref{fig:LM_PPO_2wh_bicycle} compare the performance of RPPO and PPO for various LMs respectively, on the same scenario \texttt{Repairing a Flat Bicycle Tire}. 
We also compare the performance of PPO for another scenario \texttt{Taking a Bath} in Fig. \ref{fig:LM_PPO_2wh_bath} and observe similar trends.
We also observe that among various LMs, the models trained with paraphrases perform better than the ones trained on sentences. We speculate that since the choices in the games have paraphrases like ``pick up a dish," the embeddings generated by the paraphrases ones are a better choice for the environment.


\subsection{Generalization among Scenarios}\label{app:generalization-among-scenarios}
\begin{figure}[t]
\centering
  \includegraphics[width=0.8\linewidth]{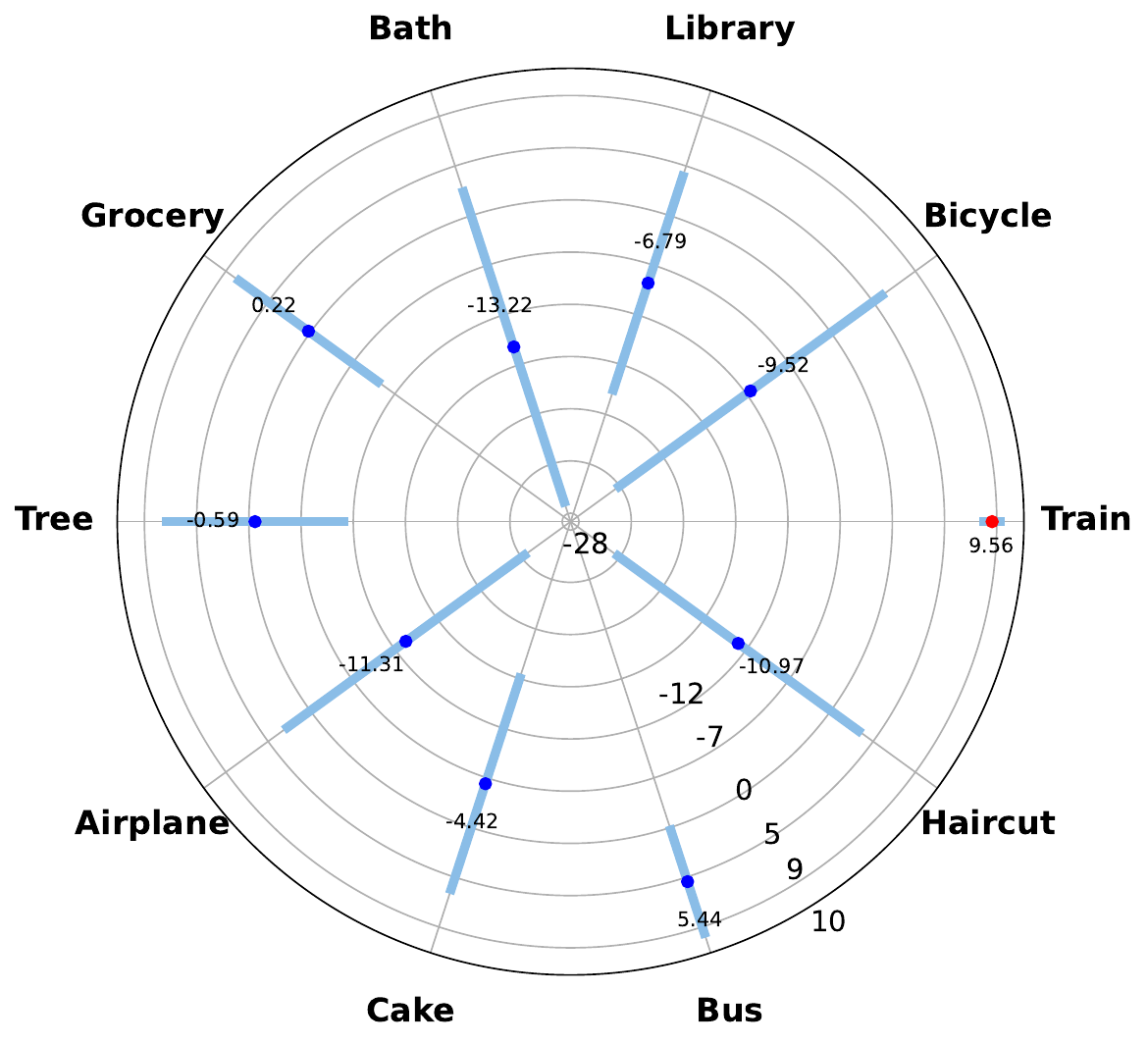}
  \caption[Generalization Performance ($\mu, \sigma$)]{{The figure shows the performance of RPPO algorithm trained on scenario \texttt{Going on a Train} on other scenarios with setting (with handicap, choices=2).}}
  \label{fig:RPPO_train_vs_all_error_plot}
\end{figure}

\begin{figure}[t]
\centering
  \includegraphics[width=\linewidth]{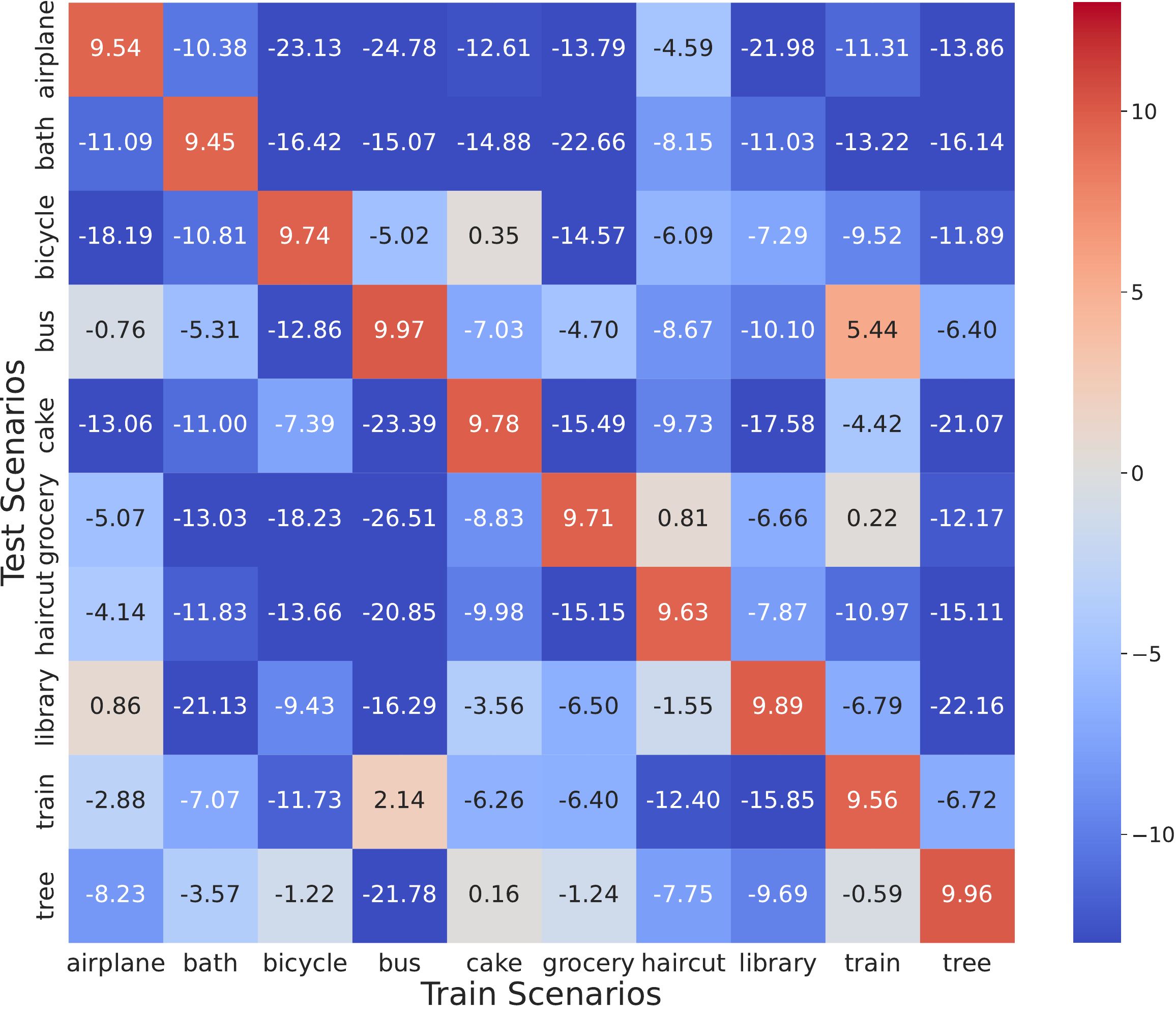}
  \caption[Heatmap RPPO (handicap, choices=2)]{{The figure shows the performance of RPPO algorithm on setting (with handicap, choices=2), trained on one scenario and tested on all scenarios.}}
  \label{fig:hm_2wh_RPPO}
\end{figure}

\begin{figure}[t]
\centering
  \includegraphics[width=\linewidth]{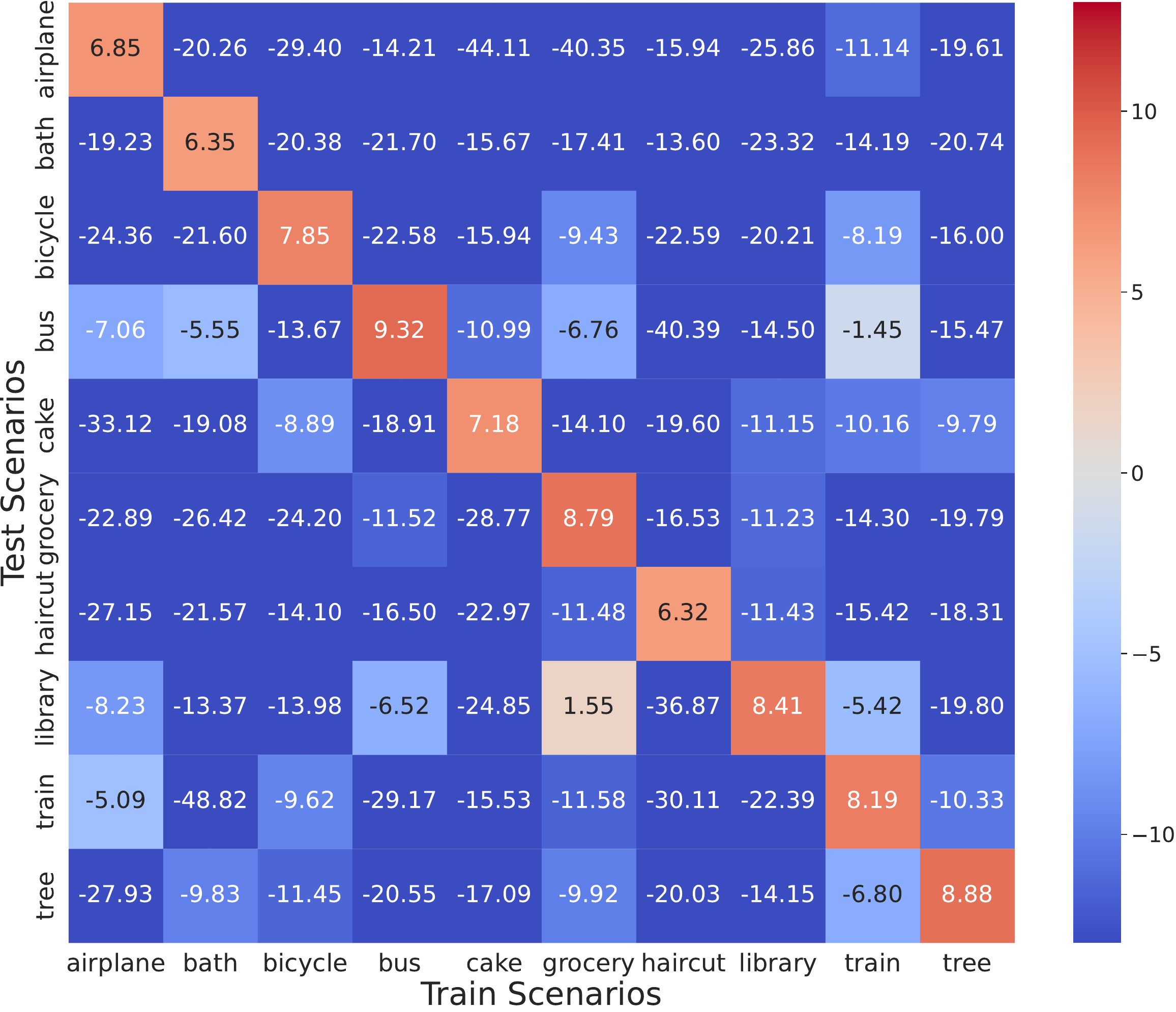}
  \caption[Heatmap RPPO (w/o handicap, choices=2)]{{The figure shows the performance of RPPO algorithm on setting (without handicap, choices=2), trained on one scenario and tested on all scenarios.}}
  \label{fig:hm_2woh_RPPO}
\end{figure}

\begin{figure}[t]
\centering
  \includegraphics[width=\linewidth]{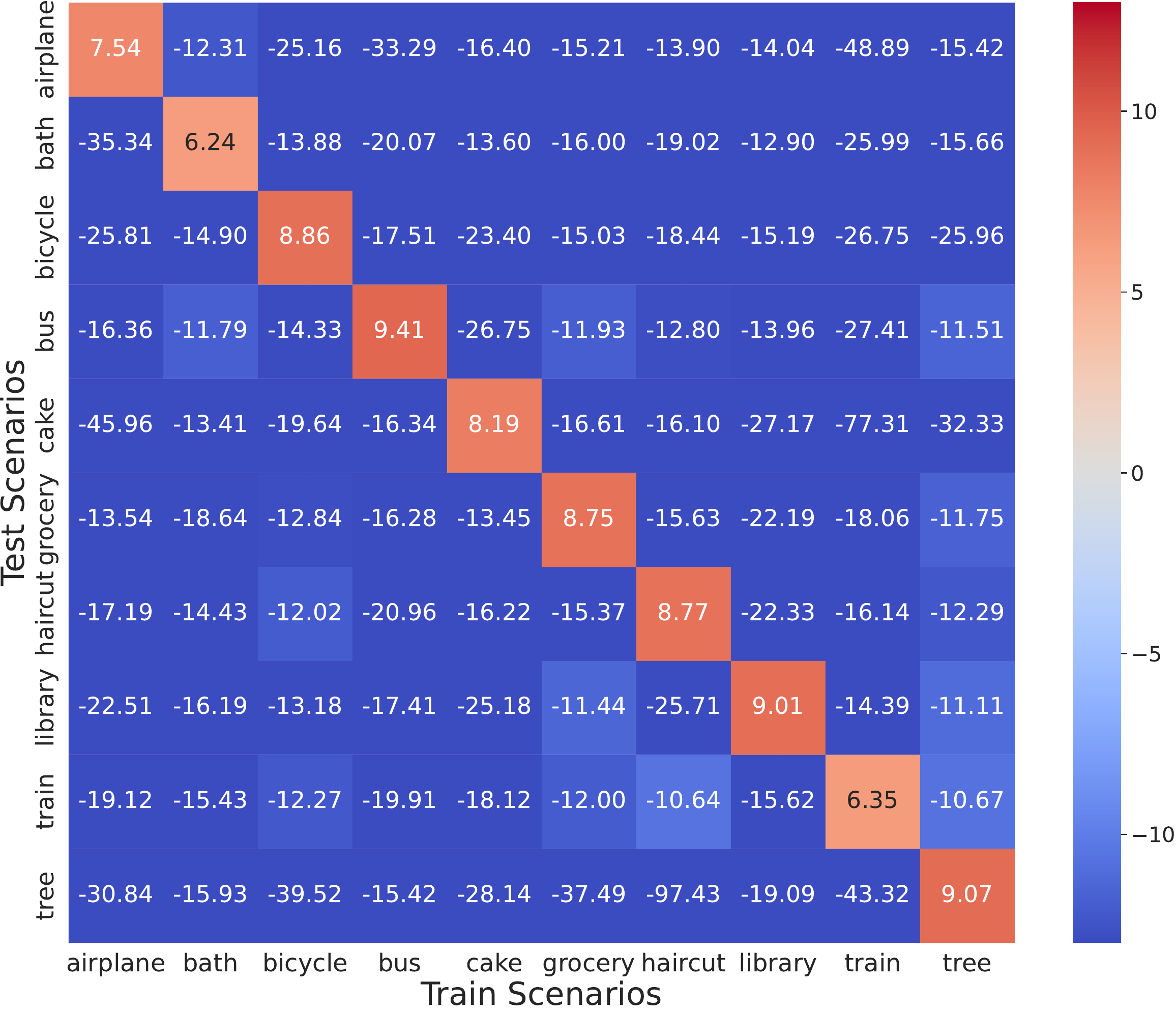}
  \caption[Heatmap RPPO (handicap, choices=5)]{{The figure shows the performance of RPPO algorithm on setting (with handicap, choices=5), trained on one scenario and tested on all scenarios.}}
  \label{fig:hm_5wh_RPPO}
\end{figure}

\begin{figure}[t]
\centering
  \includegraphics[width=\linewidth]{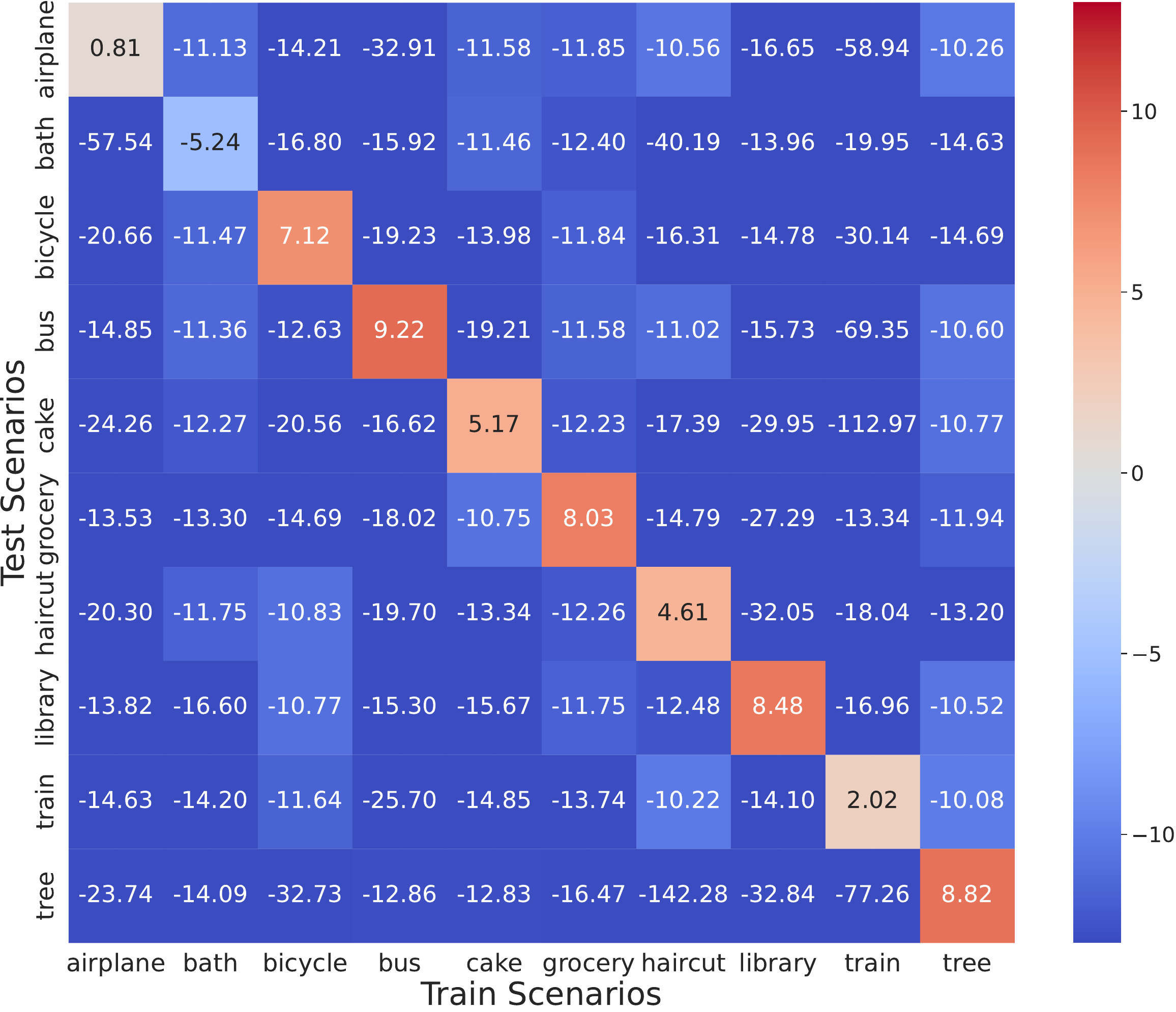}
  \caption[Heatmap RPPO (w/o handicap, choices=5)]{{The figure shows the performance of RPPO algorithm on setting (without handicap, choices=5), trained on one scenario and tested on all scenarios.}}
  \label{fig:hm_5woh_RPPO}
\end{figure}

To provide a clear idea about the generalization of an RL algorithm among the scenarios, we perform a comprehensive set of experiments by considering all possible combinations. We train the best-performing RL algorithm RPPO on all the scenarios and validate its generalization by testing the performance on all the other scenarios. As done in RL experimentations, we consider multiple seeds for testing and report the average performance over 100 test runs. Fig. \ref{fig:RPPO_train_vs_all_error_plot} shows the average performance with standard deviations plotted as an error plot. We report the generalization results on multiple settings of the environment. Fig. \ref{fig:hm_2wh_RPPO}
shows the performance of the RPPO algorithm on setting (with handicap, choices=2). We observe that the algorithms show generalizations in a few of the similar environments, like \texttt{Riding a Bus} and \texttt{Riding a Train} for this setting. However, for other difficult settings with more number choices (Fig. \ref{fig:hm_5wh_RPPO}) and without handicaps shown in Fig. \ref{fig:hm_2woh_RPPO} and Figure \ref{fig:hm_5woh_RPPO}, the generalization results are less prominent.


\vspace{50mm}
%

%

\vspace{-10pt}
%
%
%

%% file: ijcai23.bbl
\begin{thebibliography}{}

\bibitem[\protect\citeauthoryear{Adhikari \bgroup \em et al.\egroup
  }{2020}]{10.5555/3495724.3495980}
Ashutosh Adhikari, Xingdi Yuan, Marc-Alexandre C{\^o}t{\'e}, Mikulas Zelinka,
  Marc-Antoine Rondeau, Romain Laroche, Pascal Poupart, Jian Tang, Adam
  Trischler, and William~L. Hamilton.
\newblock
  \href{https://proceedings.neurips.cc/paper/2020/file/1fc30b9d4319760b04fab735fbfed9a9-Paper.pdf}{{Learning
  Dynamic Belief Graphs to Generalize on Text-Based Games}}.
\newblock In {\em NeurIPS}, 2020.

\bibitem[\protect\citeauthoryear{Adolphs and
  Hofmann}{2020}]{Adolphs2020LeDeepChefDR}
Leonard Adolphs and Thomas Hofmann.
\newblock
  \href{https://ojs.aaai.org/index.php/AAAI/article/view/6228/6084}{{LeDeepChef:
  Deep Reinforcement Learning Agent for Families of Text-Based Games}}.
\newblock In {\em AAAI}, 2020.

\bibitem[\protect\citeauthoryear{Ammanabrolu and
  Hausknecht}{2020}]{ammanabrolu2020graph}
Prithviraj Ammanabrolu and Matthew Hausknecht.
\newblock \href{https://openreview.net/forum?id=B1x6w0EtwH}{{Graph Constrained
  Reinforcement Learning for Natural Language Action Spaces}}.
\newblock In {\em International Conference on Learning Representations}, 2020.

\bibitem[\protect\citeauthoryear{Ammanabrolu and
  Riedl}{2019}]{ammanabrolu-riedl-2019-transfer}
Prithviraj Ammanabrolu and Mark Riedl.
\newblock \href{https://aclanthology.org/D19-5301}{{Transfer in Deep
  Reinforcement Learning Using Knowledge Graphs}}.
\newblock In {\em Proceedings of the Thirteenth Workshop on Graph-Based Methods
  for Natural Language Processing (TextGraphs-13)}, 2019.

\bibitem[\protect\citeauthoryear{Chatzou \bgroup \em et al.\egroup
  }{2016}]{chatzou2016multiple}
Maria Chatzou, Cedrik Magis, Jia-Ming Chang, Carsten Kemena, Giovanni Bussotti,
  Ionas Erb, and Cedric Notredame.
\newblock \href{https://doi.org/10.1093/bib/bbv099}{{Multiple Sequence
  Alignment Modeling: Methods and Applications}}.
\newblock {\em Briefings in Bioinformatics}, 2016.

\bibitem[\protect\citeauthoryear{Chaudhury \bgroup \em et al.\egroup
  }{2020}]{chaudhury2020crest}
Subhajit Chaudhury, Daiki Kimura, Kartik Talamadupula, Michiaki Tatsubori, Asim
  Munawar, and Ryuki Tachibana.
\newblock \href{https://aclanthology.org/2020.emnlp-main.241}{{Bootstrapped
  {Q}-learning with Context Relevant Observation Pruning to Generalize in
  Text-based Games}}.
\newblock In {\em Proceedings of the 2020 Conference on Empirical Methods in
  Natural Language Processing (EMNLP)}, 2020.

\bibitem[\protect\citeauthoryear{C{\^o}t{\'e} \bgroup \em et al.\egroup
  }{2018}]{cote2018textworld}
Marc-Alexandre C{\^o}t{\'e}, {\'A}kos K{\'a}d{\'a}r, Xingdi Yuan, Ben~A.
  Kybartas, Tavian Barnes, Emery Fine, James Moore, Matthew~J. Hausknecht,
  Layla~El Asri, Mahmoud Adada, Wendy Tay, and Adam Trischler.
\newblock \href{https://arxiv.org/abs/1806.11532}{{TextWorld: A Learning
  Environment for Text-based Games}}.
\newblock In {\em CGW@IJCAI}, 2018.

\bibitem[\protect\citeauthoryear{Frermann \bgroup \em et al.\egroup
  }{2014}]{frermann2014hierarchical}
Lea Frermann, Ivan Titov, and Manfred Pinkal.
\newblock \href{https://aclanthology.org/E14-1006}{{A Hierarchical {B}ayesian
  Model for Unsupervised Induction of Script Knowledge}}.
\newblock In {\em Proceedings of the 14th Conference of the {E}uropean Chapter
  of the Association for Computational Linguistics}, 2014.

\bibitem[\protect\citeauthoryear{Hausknecht \bgroup \em et al.\egroup
  }{2020}]{hausknecht2020interactive}
Matthew Hausknecht, Prithviraj Ammanabrolu, Marc-Alexandre C{\^o}t{\'e}, and
  Xingdi Yuan.
\newblock
  \href{https://ojs.aaai.org/index.php/AAAI/article/view/6297}{{Interactive
  Fiction Games: A Colossal Adventure}}.
\newblock In {\em AAAI}, 2020.

\bibitem[\protect\citeauthoryear{He \bgroup \em et al.\egroup
  }{2016}]{he-etal-2016-deep}
Ji~He, Jianshu Chen, Xiaodong He, Jianfeng Gao, Lihong Li, Li~Deng, and Mari
  Ostendorf.
\newblock \href{https://aclanthology.org/P16-1153}{{Deep Reinforcement Learning
  with a Natural Language Action Space}}.
\newblock In {\em Proceedings of the 54th Annual Meeting of the Association for
  Computational Linguistics}, 2016.

\bibitem[\protect\citeauthoryear{Hill \bgroup \em et al.\egroup
  }{2017}]{groundedLanguage2017}
Felix Hill, Karl~Moritz Hermann, Phil Blunsom, and Stephen Clark.
\newblock \href{http://arxiv.org/abs/1710.09867}{{Understanding Grounded
  Language Learning Agents}}.
\newblock {\em arXiv preprint arXiv:1710.09867}, 2017.

\bibitem[\protect\citeauthoryear{Jans \bgroup \em et al.\egroup
  }{2012}]{jans-etal-2012-skip}
Bram Jans, Steven Bethard, Ivan Vuli{\'c}, and Marie~Francine Moens.
\newblock \href{https://aclanthology.org/E12-1034}{{Skip N-grams and Ranking
  Functions for Predicting Script Events}}.
\newblock In {\em Proceedings of the 13th Conference of the {E}uropean Chapter
  of the Association for Computational Linguistics}, 2012.

\bibitem[\protect\citeauthoryear{Kaelbling \bgroup \em et al.\egroup
  }{1998}]{pomdp}
Leslie~Pack Kaelbling, Michael~L Littman, and Anthony~R Cassandra.
\newblock
  \href{https://people.csail.mit.edu/lpk/papers/aij98-pomdp.pdf}{{Planning and
  Acting in Partially Observable Stochastic Domains}}.
\newblock {\em Artificial Intelligence}, 1998.

\bibitem[\protect\citeauthoryear{K\"{u}ttler \bgroup \em et al.\egroup
  }{2020}]{Kuttler2020TheNL}
Heinrich K\"{u}ttler, Nantas Nardelli, Alexander~H. Miller, Roberta Raileanu,
  Marco Selvatici, Edward Grefenstette, and Tim Rockt\"{a}schel.
\newblock \href{https://arxiv.org/abs/2006.13760}{{The NetHack Learning
  Environment}}.
\newblock In {\em NeurIPS}, 2020.

\bibitem[\protect\citeauthoryear{Miikkulainen and
  Elman}{1993}]{miikkulainen1993subsymbolic}
Risto Miikkulainen and Jeffrey Elman.
\newblock {\em \href{https://aclanthology.org/J94-1012.pdf}{{Subsymbolic
  Natural Language Processing: An Integrated Model of Scripts, Lexicon, and
  Memory}}}.
\newblock MIT press, 1993.

\bibitem[\protect\citeauthoryear{Mnih \bgroup \em et al.\egroup
  }{2013}]{mnih2013playing}
Volodymyr Mnih, Koray Kavukcuoglu, David Silver, Alex Graves, Ioannis
  Antonoglou, Daan Wierstra, and Martin Riedmiller.
\newblock \href{https://www.cs.toronto.edu/~vmnih/docs/dqn.pdf}{{Playing Atari
  with Deep Reinforcement Learning}}.
\newblock {\em arXiv preprint arXiv:1312.5602}, 2013.

\bibitem[\protect\citeauthoryear{Mnih \bgroup \em et al.\egroup
  }{2016}]{mnih2016asynchronous_A2C_A3C}
Volodymyr Mnih, Adria~Puigdomenech Badia, Mehdi Mirza, Alex Graves, Timothy
  Lillicrap, Tim Harley, David Silver, and Koray Kavukcuoglu.
\newblock \href{https://dl.acm.org/doi/10.5555/3045390.3045594}{{Asynchronous
  Methods for Deep Reinforcement Learning}}.
\newblock In {\em {International Conference on Machine Learning}}, 2016.

\bibitem[\protect\citeauthoryear{Modi and
  Titov}{2014}]{modi-titov-2014-inducing}
Ashutosh Modi and Ivan Titov.
\newblock \href{https://aclanthology.org/W14-1606}{{Inducing Neural Models of
  Script Knowledge}}.
\newblock In {\em Proceedings of the Eighteenth Conference on Computational
  Natural Language Learning}, 2014.

\bibitem[\protect\citeauthoryear{Modi \bgroup \em et al.\egroup
  }{2016}]{modi-etal-2016-inscript}
Ashutosh Modi, Tatjana Anikina, Simon Ostermann, and Manfred Pinkal.
\newblock \href{https://aclanthology.org/L16-1555}{{{I}n{S}cript: Narrative
  texts annotated with script information}}.
\newblock In {\em Proceedings of the Tenth International Conference on Language
  Resources and Evaluation ({LREC}'16)}, 2016.

\bibitem[\protect\citeauthoryear{Modi \bgroup \em et al.\egroup
  }{2017}]{modi-etal-2017-modeling}
Ashutosh Modi, Ivan Titov, Vera Demberg, Asad Sayeed, and Manfred Pinkal.
\newblock \href{https://aclanthology.org/Q17-1003}{{Modeling Semantic
  Expectation: Using Script Knowledge for Referent Prediction}}.
\newblock {\em Transactions of the Association for Computational Linguistics},
  2017.

\bibitem[\protect\citeauthoryear{Modi}{2016}]{modi-2016-event}
Ashutosh Modi.
\newblock \href{https://aclanthology.org/K16-1008}{{Event Embeddings for
  Semantic Script Modeling}}.
\newblock In {\em Proceedings of the 20th {SIGNLL} Conference on Computational
  Natural Language Learning}, 2016.

\bibitem[\protect\citeauthoryear{Modi}{2017}]{Modi17-Thesis}
Ashutosh Modi.
\newblock {\em \href{https://d-nb.info/1137206276/34}{Modeling Common Sense
  Knowledge via Scripts}}.
\newblock PhD thesis, Saarland University, 2017.

\bibitem[\protect\citeauthoryear{Murugesan \bgroup \em et al.\egroup
  }{2020}]{Murugesan2021TextbasedRA}
Keerthiram Murugesan, Mattia Atzeni, Pavan Kapanipathi, Pushkar Shukla, Sadhana
  Kumaravel, Gerald Tesauro, Kartik Talamadupula, Mrinmaya Sachan, and Murray
  Campbell.
\newblock
  \href{https://ojs.aaai.org/index.php/AAAI/article/view/17090/16897}{{Text-based
  RL Agents with Commonsense Knowledge: New Challenges, Environments and
  Baselines}}.
\newblock In {\em AAAI Conference on Artificial Intelligence}, 2020.

\bibitem[\protect\citeauthoryear{Narasimhan \bgroup \em et al.\egroup
  }{2015}]{narasimhan2015language}
Karthik Narasimhan, Tejas Kulkarni, and Regina Barzilay.
\newblock \href{https://aclanthology.org/D15-1001}{{Language Understanding for
  Text-based Games using Deep Reinforcement Learning}}.
\newblock In {\em Proceedings of the 2015 Conference on Empirical Methods in
  Natural Language Processing}, 2015.

\bibitem[\protect\citeauthoryear{Nguyen \bgroup \em et al.\egroup
  }{2019}]{nguyen2019toward}
Cuong~V Nguyen, Alessandro Achille, Michael Lam, Tal Hassner, Vijay Mahadevan,
  and Stefano Soatto.
\newblock \href{https://arxiv.org/abs/1908.01091}{{Toward Understanding
  Catastrophic Forgetting in Continual Learning}}.
\newblock {\em arXiv preprint arXiv:1908.01091}, abs/1908.01091, 2019.

\bibitem[\protect\citeauthoryear{Ostermann \bgroup \em et al.\egroup
  }{2018a}]{ostermann-etal-2018-mcscript}
Simon Ostermann, Ashutosh Modi, Michael Roth, Stefan Thater, and Manfred
  Pinkal.
\newblock \href{https://aclanthology.org/L18-1564}{{{MCS}cript: A Novel Dataset
  for Assessing Machine Comprehension Using Script Knowledge}}.
\newblock In {\em {LREC}}, 2018.

\bibitem[\protect\citeauthoryear{Ostermann \bgroup \em et al.\egroup
  }{2018b}]{ostermann-etal-2018-semeval}
Simon Ostermann, Michael Roth, Ashutosh Modi, Stefan Thater, and Manfred
  Pinkal.
\newblock \href{https://aclanthology.org/S18-1119}{{{S}em{E}val-2018 Task 11:
  Machine Comprehension Using Commonsense Knowledge}}.
\newblock In {\em Proceedings of the 12th International Workshop on Semantic
  Evaluation}, 2018.

\bibitem[\protect\citeauthoryear{Pennington \bgroup \em et al.\egroup
  }{2014}]{pennington2014glove}
Jeffrey Pennington, Richard Socher, and Christopher Manning.
\newblock \href{https://aclanthology.org/D14-1162}{{{G}lo{V}e: Global Vectors
  for Word Representation}}.
\newblock In {\em Proceedings of the 2014 Conference on Empirical Methods in
  Natural Language Processing ({EMNLP})}, 2014.

\bibitem[\protect\citeauthoryear{Pichotta and
  Mooney}{2016}]{pichotta-mooney-2016-using}
Karl Pichotta and Raymond~J. Mooney.
\newblock \href{https://aclanthology.org/P16-1027}{{Using Sentence-Level {LSTM}
  Language Models for Script Inference}}.
\newblock In {\em Proceedings of the 54th Annual Meeting of the Association for
  Computational Linguistics}, 2016.

\bibitem[\protect\citeauthoryear{Radford \bgroup \em et al.\egroup
  }{2019}]{radford2019language}
Alec Radford, Jeffrey Wu, Rewon Child, David Luan, Dario Amodei, Ilya
  Sutskever, et~al.
\newblock Language models are unsupervised multitask learners.
\newblock {\em OpenAI blog}, 1(8):9, 2019.

\bibitem[\protect\citeauthoryear{Regneri \bgroup \em et al.\egroup
  }{2010}]{regneri2010learning}
Michaela Regneri, Alexander Koller, and Manfred Pinkal.
\newblock \href{https://aclanthology.org/P10-1100}{{Learning Script Knowledge
  with Web Experiments}}.
\newblock In {\em Proceedings of the 48th Annual Meeting of the Association for
  Computational Linguistics}, 2010.

\bibitem[\protect\citeauthoryear{Reimers and
  Gurevych}{2019}]{reimers-2019-sentence-bert-sbert}
Nils Reimers and Iryna Gurevych.
\newblock \href{https://aclanthology.org/D19-1410}{{Sentence-{BERT}: Sentence
  Embeddings using {S}iamese {BERT}-Networks}}.
\newblock In {\em Proceedings of the 2019 Conference on Empirical Methods in
  Natural Language Processing and the 9th International Joint Conference on
  Natural Language Processing (EMNLP-IJCNLP)}, 2019.

\bibitem[\protect\citeauthoryear{Rudinger \bgroup \em et al.\egroup
  }{2015}]{rudinger-etal-2015-learning}
Rachel Rudinger, Vera Demberg, Ashutosh Modi, Benjamin Van~Durme, and Manfred
  Pinkal.
\newblock \href{https://aclanthology.org/S15-1024}{{Learning to predict script
  events from domain-specific text}}.
\newblock In {\em Proceedings of the Fourth Joint Conference on Lexical and
  Computational Semantics}, 2015.

\bibitem[\protect\citeauthoryear{Sakaguchi \bgroup \em et al.\egroup
  }{2021}]{sakaguchi-etal-2021-proscript-partially}
Keisuke Sakaguchi, Chandra Bhagavatula, Ronan Le~Bras, Niket Tandon, Peter
  Clark, and Yejin Choi.
\newblock \href{https://aclanthology.org/2021.findings-emnlp.184}{{pro{S}cript:
  Partially Ordered Scripts Generation}}.
\newblock In {\em Findings of the Association for Computational Linguistics:
  EMNLP}, 2021.

\bibitem[\protect\citeauthoryear{Sancheti and Rudinger}{2022}]{lm-scripts-2021}
Abhilasha Sancheti and Rachel Rudinger.
\newblock \href{https://aclanthology.org/2022.starsem-1.1}{{What do Large
  Language Models Learn about Scripts?}}
\newblock In {\em Proceedings of the 11th Joint Conference on Lexical and
  Computational Semantics}, 2022.

\bibitem[\protect\citeauthoryear{Schank and Abelson}{1975}]{schank1975scripts}
Roger~C. Schank and Robert~P. Abelson.
\newblock \href{https://dl.acm.org/doi/abs/10.5555/1624626.1624649}{{Scripts,
  Plans, and Knowledge}}.
\newblock In {\em Proceedings of the 4th International Joint Conference on
  Artificial Intelligence}, IJCAI, 1975.

\bibitem[\protect\citeauthoryear{Schulman \bgroup \em et al.\egroup
  }{2017}]{PPO}
John Schulman, Filip Wolski, Prafulla Dhariwal, Alec Radford, and Oleg Klimov.
\newblock Proximal policy optimization algorithms.
\newblock {\em arXiv preprint arXiv:1707.06347}, 2017.

\bibitem[\protect\citeauthoryear{Singh \bgroup \em et al.\egroup
  }{2022}]{Singh2022PretrainedLM}
Ishika Singh, Gargi Singh, and Ashutosh Modi.
\newblock
  \href{https://www.ifaamas.org/Proceedings/aamas2022/pdfs/p1729.pdf}{{Pre-trained
  Language Models as Prior Knowledge for Playing Text-based Games}}.
\newblock In {\em 21st International Conference on Autonomous Agents and
  Multiagent Systems, {AAMAS}}, 2022.

\bibitem[\protect\citeauthoryear{Sutton and
  Barto}{2018}]{sutton2018reinforcement}
Richard~S Sutton and Andrew~G Barto.
\newblock {\em
  \href{http://incompleteideas.net/book/RLbook2020.pdf}{Reinforcement Learning:
  An Introduction}}.
\newblock MIT press, 2018.

\bibitem[\protect\citeauthoryear{Wanzare \bgroup \em et al.\egroup
  }{2016}]{wanzare-etal-2016-crowdsourced}
Lilian D.~A. Wanzare, Alessandra Zarcone, Stefan Thater, and Manfred Pinkal.
\newblock \href{https://aclanthology.org/L16-1556}{{A Crowdsourced Database of
  Event Sequence Descriptions for the Acquisition of High-quality Script
  Knowledge}}.
\newblock In {\em Proceedings of the Tenth International Conference on Language
  Resources and Evaluation ({LREC}'16)}, 2016.

\bibitem[\protect\citeauthoryear{Yao \bgroup \em et al.\egroup
  }{2020}]{Yao2020KeepCA}
Shunyu Yao, Rohan Rao, Matthew Hausknecht, and Karthik Narasimhan.
\newblock \href{https://aclanthology.org/2020.emnlp-main.704}{{Keep {CALM} and
  Explore: Language Models for Action Generation in Text-based Games}}.
\newblock In {\em Proceedings of the 2020 Conference on Empirical Methods in
  Natural Language Processing (EMNLP)}, 2020.

\bibitem[\protect\citeauthoryear{Yin and May}{2019}]{Yin2019LearnHT}
Xusen Yin and Jonathan May.
\newblock \href{https://arxiv.org/abs/1908.04777}{{Learn how to cook a new
  recipe in a new house: Using map familiarization, curriculum learning, and
  bandit feedback to learn families of text-based adventure games}}.
\newblock {\em arXiv preprint arXiv:1908.04777}, 2019.

\bibitem[\protect\citeauthoryear{Yuan \bgroup \em et al.\egroup
  }{2019}]{Yuan2019InteractiveLL}
Xingdi Yuan, Marc-Alexandre C{\^o}t{\'e}, Jie Fu, Zhouhan Lin, Chris Pal,
  Yoshua Bengio, and Adam Trischler.
\newblock \href{https://aclanthology.org/D19-1280}{{Interactive Language
  Learning by Question Answering}}.
\newblock In {\em Proceedings of the 2019 Conference on Empirical Methods in
  Natural Language Processing and the 9th International Joint Conference on
  Natural Language Processing (EMNLP-IJCNLP)}, 2019.

\end{thebibliography}
